\documentclass{interact}

\usepackage{epstopdf}
\usepackage{xcolor}

\usepackage{natbib}
\bibpunct[, ]{(}{)}{;}{a}{}{,}

\theoremstyle{plain}

\theoremstyle{definition}

\theoremstyle{remark}

\usepackage[T1]{fontenc}

\usepackage[hidelinks, colorlinks = true,
            linkcolor = blue,
            urlcolor  = blue,
            citecolor = blue,
            anchorcolor = blue]{hyperref}
\usepackage[none]{hyphenat}
\usepackage{pifont}
\newcommand{\cmark}{\ding{51}}
\newcommand{\xmark}{\ding{55}}
\usepackage{subfig}

\usepackage{algorithm}
\usepackage[noend]{algpseudocode}
\usepackage{float}

\usepackage{wrapfig}
\usepackage[super]{nth}
\usepackage{mathtools}
\usepackage[c2]{optidef}
\usepackage[labelformat=simple]{caption}
\usepackage{amsmath}

\newcommand*\sqcitep[1]{{\setcitestyle{square}$\!\!$\citep{#1}}}
\usepackage{listings}
\usepackage{xcolor}

\newcommand{\gray}[1]{\textcolor{codegray}{#1}}
\definecolor{codegray}{rgb}{0.5,0.5,0.5}

\begin{document}


\title{Fair Collaborative Vehicle Routing: A Deep Multi-Agent\\Reinforcement Learning Approach}

\author{
    \name{Stephen Mak\textsuperscript{a,$\ast$}\thanks{$^\ast$Corresponding author: sm2410@cam.ac.uk, 17 Charles Babbage Road, CB3 0FS, United Kingdom},
    Liming Xu\textsuperscript{a},
    Tim Pearce\textsuperscript{b,1}\thanks{\textsuperscript{1}Previously at Department of Computer Science and Technology, Tsinghua University},
    Michael Ostroumov\textsuperscript{c},\\
    Alexandra Brintrup\textsuperscript{a}}
    \affil{
    \textsuperscript{a}Institute for Manufacturing, Department of Engineering, University of Cambridge\\
    \textsuperscript{b}Microsoft Research Cambridge\\ 
    \textsuperscript{c}Value Chain Lab}
}

\maketitle

\begin{abstract}

Collaborative vehicle routing occurs when carriers collaborate through sharing their transportation requests and performing transportation requests on behalf of each other. This achieves economies of scale, thus reducing cost, greenhouse gas emissions and road congestion. But which carrier should partner with whom, and how much should each carrier be compensated? Traditional game theoretic solution concepts are expensive to calculate as the characteristic function scales exponentially with the number of agents. This would require solving the vehicle routing problem (NP-hard) an exponential number of times. We therefore propose to model this problem as a coalitional bargaining game solved using deep multi-agent reinforcement learning, where – crucially – agents are \emph{not} given access to the characteristic function. Instead, we \emph{implicitly} reason about the characteristic function; thus, when deployed in production, we only need to evaluate the expensive post-collaboration vehicle routing problem once. Our contribution is that we are the first to consider both the route allocation problem and gain sharing problem simultaneously \--- without access to the expensive characteristic function. Through decentralised machine learning, our agents bargain with each other and agree to outcomes that correlate well with the Shapley value \--- a fair profit allocation mechanism. Importantly, we are able to achieve a reduction in run-time of 88\%.

\end{abstract}


\section{Introduction}

Heavy goods vehicles (HGVs) in the UK contributed 4.3\% of the UK’s \emph{total} greenhouse gas emissions in 2019 \citep{uk_beis_final_2021}. HGVs are utilised inefficiently at 61\% of their total weight capacity. Moreover, 30\% of the distance travelled carries zero freight \citep[RFS0125]{uk_dft_road_2020}.

\begin{figure}
  \captionsetup[subfigure]{justification=centering}
  \centering
  \subfloat[Pre-collaboration (total cost: 3.35)]{
    \includegraphics[width=0.445\linewidth]{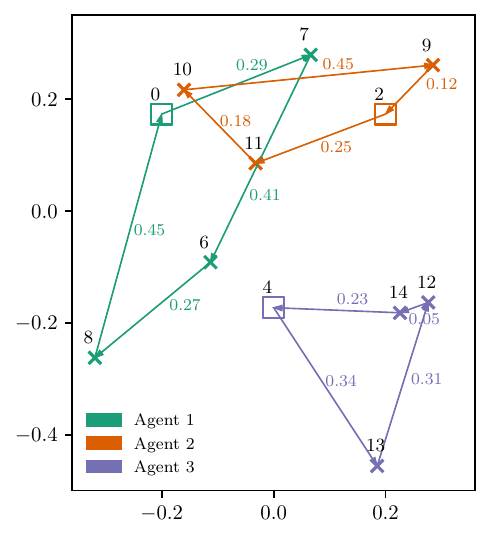}\label{fig: f1a}
  }
  \subfloat[Post-collaboration with coalition $\{1, 2, 3 \}$ (total cost: 2.47)]{
    \includegraphics[width=0.445\linewidth]{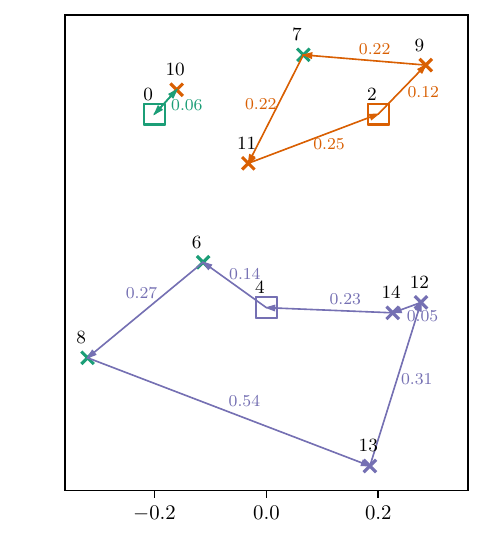}\label{fig: f1b}
  }
  \caption{
  Three agents (denoted by colours) before and after collaboration. Squares denote depots. Crosses denote customer locations. Node indices (arbitrary) are denoted in black, with costs given in their respective colours. The \emph{collaboration gain} is defined as the difference in social welfare (or total cost) before and after collaboration. In \autoref{fig: f1b}, Agents 1, 2 and 3 all decide to collaborate which reduces the system's total cost by 0.88 (or 26\%). This results in a \emph{collaboration gain per capita} (assuming agents split the gain equally) of 0.29. For detailed calculations, see \autoref{sec: collaborative_vehicle_routing}.
  }
  \label{fig: f1}
\end{figure}

Collaborative vehicle routing (CVR) has been proposed to improve HGV utilisation. Here, carriers collaborate through sharing their delivery information in order to achieve economies of scale. If carriers agree to work together, they are said to be in a \emph{coalition}. As a result of improved utilisation, total travel costs across collaborating carriers can be reduced, resulting in a \emph{collaboration gain}. The remaining question then is how to allocate this collaboration gain in a fair manner such that carriers are incentivised to form coalitions. An example of CVR is given in \autoref{fig: f1}.

Prior literature suggests that collaborative vehicle
routing can reduce costs by around 4\---46\% and also reduce greenhouse gas emissions and road congestion \citep{cruijssen_joint_2007, zhang_optimal_2017, gansterer_collaborative_2018, pan_horizontal_2019, gansterer_shared_2020, cruijssen_cross-chain_2020, ferrell_horizontal_2020}. Sharing resources may also lead to improved resilience to fluctuations in supply and/or demand. Despite these benefits, real-world adoption remains limited, with only a few companies participating \citep{cruijssen_horizontal_2007, guajardo_review_2016, cruijssen_cross-chain_2020}. Currently, a key barrier is the computational complexity of calculating a fair \emph{gain sharing} mechanism that scales with a larger number of companies. \citet{guajardo_review_2016} recommends that future work should investigate approximate gain sharing methods. Our paper follows this recommendation.

Our first contribution is modelling the collaborative routing problem as a \emph{coalitional bargaining game} \citep{okada_noncooperative_1996} with intelligent agents obtained through the use of \emph{deep multi-agent reinforcement learning} (MARL). We provide the theoretical grounding in this paper, tying together the fields of collaborative vehicle routing, coalitional bargaining, and deep multi-agent reinforcement learning in order to obtain a theoretically grounded approach that significantly reduces run-time. Here, agents attempt to reach agreement on selecting the `best' carrier(s) to partner with, and  rationally share the collaboration gain amongst the coalition. This bargaining process takes place over multiple rounds of bargaining (see \autoref{sec: coalitional_bargaining} for a formal definition). A benefit of this approach is that both the routing problem (who should deliver which requests?) and the gain sharing problem (who receives how much of the added value?) are considered simultaneously, whereas a key limitation of many previous methods consider these sub-problems in isolation from one another \citep{gansterer_collaborative_2018}. Moreover, our approach is agnostic to the underlying routing problem \--- the complexity of the vehicle routing problem (VRP) formulation can be increased with further constraints such as  time windows, without further modification to the method.

Our second contribution is that agents do not need access to the full characteristic function explicitly. To obtain the full characteristic function, the collaboration gain for all possible coalitions must be calculated. In the three-player setting, there are four possible coalitions $\{1, 2, 3\}, \{1, 2\}, \{1, 3\}$ and $ \{2, 3\} $. Therefore, to obtain the full characteristic function requires solving $2^{n-1}$ NP-hard post-collaboration VRPs (for a formal introduction, see \autoref{sec: gain_sharing}. As a result, methods that require full access to the characteristic function are intractable for settings with more than 6 carriers \citep{cruijssen_cross-chain_2020}. Instead, our agents can implicitly reason about the characteristic function through only receiving a high-dimensional graph input of delivery information (for example, latitudes and longitudes), as well as other agents' actions. This eliminates the need to fully evaluate the characteristic function when deployed in production, which involves solving the expensive post-collaboration VRP an exponential number of times. Instead, we only need to solve the post-collaboration VRP once when deployed in real-world settings, thus allowing our approach to achieve a significant run-time reduction. In addition, our approach utilises Centralised Training with Decentralised Execution (CTDE) to obtain decentralised agent policies \citep{lowe_multi-agent_2020}. Decentralised policies are desirable in real-world applications as each agent does not necessarily require access to the global, underlying state. This helps ensure that companies' sensitive information will not be leaked to competitors. This also aids to stabilise training in multi-agent settings as well as reduce communication costs. Furthermore, our approach is \emph{inductive} as opposed to transductive of prior methods. This enables our agents to generalise to agents never seen before during training and thus reduces computational cost.

The remainder of this paper is organised as follows. \autoref{sec: related_work} positions our work within the wider context of both collaborative vehicle routing and deep multi-agent reinforcement learning. \autoref{sec: background} provides a formal introduction to coalitional games, coalitional bargaining and reinforcement learning. \autoref{sec: methodology} discusses and justifies various design decisions regarding our agents. \autoref{sec: exp} details our experimental setup, results, discussion and future work. Finally, \autoref{sec: conclusion} concludes our findings and provides broader managerial implications as a result of this work.

\section{Related Work}
\label{sec: related_work}

\subsection{Collaborative vehicle routing}
Prior collaborative routing literature tackles the partner selection sub-problem (i.e., who should each carrier work with?) by estimating the collaboration gain between different carriers using heuristics \citep{palhazi_cuervo_determining_2016, adenso-diaz_analysis_2014}. However, a limitation of this approach is that they do not consider how much each agent should be compensated, nor if agents even agree to join the same coalitions (i.e., if the coalitions are stable). Posing this problem as a \emph{coalitional bargaining} game not only allows us to tackle the partner selection aspect, but we are also able to consider the gain sharing aspect simultaneously as well.

The majority of the collaborative routing literature is concerned with the exchange of \emph{individual} transportation requests amongst the carriers. This can be divided into three types of planning approaches: centralised; decentralised without auctions; and decentralised with auctions \citep{gansterer_collaborative_2018, gansterer_shared_2020}.

\subsubsection{Centralised planning}
Centralised planning approaches desire to simply maximise social welfare (the sum of each company's profits). Typically, this goal is achieved by using a form of mixed integer linear programming or (meta)heuristics \citep{cruijssen_joint_2007, gansterer_collaborative_2018, angelelli_optimization_2022}. This can be viewed as a common-payoff setting, i.e., where all agents are on the same team and receive the same reward.  However, assuming a common-payoff setting in practice is unrealistic as companies are \emph{self-interested} \--- they mostly care only about their own profits \citep{cruijssen_horizontal_2007}. Moreover, there exists fierce competition especially in horizontal collaborations. Therefore, the more realistic setting of decentralised control is needed where agents are modelled to be self-interested.

\subsubsection{Decentralised planning}
There have been few attempts to tackle CVR with decentralised approaches as well. One approach focuses on the problem of \emph{partner selection}, i.e. ``who should work with whom?''. \citet{adenso-diaz_analysis_2014} proposes an a priori index to estimate the collaboration gain between carriers based on their transportation requests. However, a key limitation is that they do not consider the gain sharing aspect and thus the coalitions formed may not be stable.

A key challenge in decentralised settings is managing the explosion in the number of \emph{bundles}. Consider \autoref{fig: f1} where Agent 2 may desire to sell delivery node 10 to Agent 1. However, if Agent 2 offers both nodes 10 and 11 as a \emph{bundle}, then Agent 2 may be able to command a higher price. Indeed, the number of possible bundles scales $\mathcal{O}(2^m)$ where $m$ is the number of deliveries. To manage this explosion, a heuristic is typically implemented where agents can only submit or request a few bundles (sometimes only one) which would limit optimality \citep{bo_dai_mathematical_2009}.

A second challenge is to also elicit other agents' preferences over all bundles. One approach is to invoke structure on the problem in the form of \emph{combinatorial auctions} which aids optimality \citep{krajewska_horizontal_2008, gansterer_collaborative_2018, gansterer_cost_2019, los_large-scale_2022}. Auctions are where carriers submit requests they do not wish to fulfil to a common pool. Then, other carriers can submit bids on these requests with various methods of determining the “winners” of said bids. Combinatorial auctions in these settings allow carriers to bid on bundles of transportation requests instead of individual transportation requests which increases its expressivity and optimality. However, this additional structure comes at additional computational complexity. Moreover, in auction mechanism design, there are four desirable properties: efficiency; individual rationality; incentive compatibility; and budget balance. \citep{gansterer_cost_2019} proposes two auction-based approaches which may be useful in practice, but would be unable to satisfy all four properties simultaneously: there exists a trade-off instead. \citep{los_large-scale_2022} investigates large-scale carrier collaboration containing 1,000 carriers with decentralised auctions. Whilst impressive in scale, their approach ignores the difficulty of large-scale gain sharing.

Both auction-based and non-auction-based approaches may also be exploited by strategic agent behaviour. Would agents intentionally misreport the costs associated with performing deliveries in order to maximise their own profits? Whilst we do not tackle this problem in our work, we believe MARL could be a useful tool to investigate this strategic behaviour in future work.

\subsubsection{Gain sharing}
\label{sec: gain_sharing}

Whilst gain sharing has been studied in collaborative routing using cooperative game theory \citep{guajardo_review_2016}, the solution concepts typically assumes that the characteristic function is given. For a set of $n$ agents, $N = \{1, \dots, n\}$, the characteristic function $v: \mathbf{2}^N \to \mathbb{R}_{\geq0}$ assigns a \emph{value}, or in our case \emph{collaboration gain}, for every possible coalition that could be formed. Note that there exists $\mathcal{O}(2^n)$ possible coalitions. This is intractable for settings with more than a few agents, because evaluating the collaboration gain of even a single coalition, involves solving a vehicle routing problem which is NP-hard. For detailed calculations of the collaboration gain, see \autoref{sec: collaborative_vehicle_routing}. \citet{guajardo_review_2016} reviews 55 papers from the collaborative transportation literature concerning gain sharing. They recommend that a future research direction should focus on developing approximate gain sharing approaches based on cooperative game theory that scales with the number of agents.

In the wider algorithmic game theory literature, coalition formation has also been extensively studied \citep{chalkiadakis_computational_2011}. However, much of the existing literature again assumes that the full characteristic function is given. Alternatively, they aim to find more succinct representations of the characteristic function, typically at a cost of increased computational complexity when computing solution concepts \citep{chalkiadakis_computational_2011}. Examples include Induced Subgraph Games and Marginal Contribution Nets \citep{deng_complexity_1994, ieong_marginal_2005}; however, even these succinct representation schemes require evaluating the value of multiple coalitions and thus solving multiple NP-hard VRPs. We argue that many real-world scenarios consist of the characteristic function being a function of the agents' assets or capabilities. In the collaborative routing setting, this is a function of the transportation requests an agent possesses. We therefore ask:\emph{ ``Can agents form optimal coalitions from the delivery information alone instead of having full access to the characteristic function?''}. Therefore, our paper can be viewed as using an alternative, succinct representation scheme which approximates a rational outcome by using a function approximator.

\subsection{Deep multi-agent reinforcement learning}
\begin{table}
    \tbl{Characteristics of selected games studied in MARL.}
    {\begin{tabular}{lcccc} \toprule
        
        Game & $>$ 2-players & Mixed-Motive & \shortstack{Known \\Optimum} & \shortstack{Partially \\Observable} \\ \midrule
        
        Go \citep{silver_mastering_2016} & \xmark & \xmark & \xmark & \xmark \\
        
        StarCraft \MakeUppercase{\romannumeral 2} \citep{vinyals_grandmaster_2019} & \xmark & \xmark & \xmark & \cmark \\
        
        SMAC\textsuperscript{a} \citep{samvelyan_starcraft_2019} & \cmark & \xmark & \xmark & \cmark \\
        
        Dota 2 \citep{openai_dota_2019} & \cmark & \xmark & \xmark & \cmark \\
    
        Gran Turismo \citep{wurman_outracing_2022} & \cmark & \xmark & \xmark & \cmark\\
    
        Football \citep{kurach_google_2020} & \cmark & \xmark & \xmark & \hphantom{d}\cmark \textsuperscript{d} \\
    
        Hide and Seek \citep{baker_emergent_2020} & \cmark & \xmark & \xmark & \cmark \\
    
        Communication \citep{foerster_learning_2016} & \cmark & \xmark & \cmark & \cmark \\
    
        GCE\textsuperscript{b} \citep{mordatch_emergence_2018} & \cmark & \xmark & \cmark & \cmark \\
    
        SSDs\textsuperscript{c} \citep{leibo_multi-agent_2017} & \cmark & \cmark & \xmark & \cmark \\
    
        {\bf Coalitional Bargaining (ours)} & \cmark & \cmark & \cmark & \xmark\\ \bottomrule
        
    \end{tabular}}
    \tabnote{
        \textsuperscript{a}StarCraft Multi-Agent Challenge; \textsuperscript{b}Grounded Communication Environment; \textsuperscript{c}Sequential Social Dilemmas; 
        \textsuperscript{d}Both fully and partially observable settings supported.
    }
    \label{table_1}
\end{table}

Single agent reinforcement learning has seen increasing adoption in supply chain management. However, supply chains can be naturally modelled as a system comprising multiple self-interested agents \citep{fox_agent-oriented_2000, xu_will_2021, brintrup_ai_2021}. For a thorough review of reinforcement learning applied towards supply chain management, see \citet{yan_reinforcement_2022}.

Recently, MARL has seen success in playing board and video games such as Go, StarCraft \MakeUppercase{\romannumeral 2} and Dota 2 \citep{silver_mastering_2016, vinyals_grandmaster_2019, openai_dota_2019}. Whilst these are tremendous feats in the AI space, the underlying games tend to be 2-player and zero-sum. However, most real-world applications, including supply chain management \citep{gabel_distributed_2012, kosasih_reinforcement_2021}, are $n$-player and mixed-motive (with potential `sequential social dilemmas' \sqcitep{leibo_multi-agent_2017}). Whilst there is some research in this direction, the majority of MARL research focuses on pure coordination or pure competition settings (see \autoref{table_1}). Our work is 3-player and mixed-motive which leads to a more challenging joint-policy space, allowing for complex behaviours such as collusion. 

The most similar work to ours from a multi-agent learning perspective is that of \citet{bachrach_negotiating_2020} and \citet{chalkiadakis_bayesian_2004}. In \citet{bachrach_negotiating_2020}, they apply deep MARL to a spatial and non-spatial Weighted Voting Game, where agents are given full access to the characteristic function. In \citet{chalkiadakis_bayesian_2004}, they apply a Bayesian MARL approach to coalition formation as their problem has uncertainty in the characteristic function. In their problem, each agent knows its own capability, but does not observe other agents' capabilities. As a result, they maintain a belief over other agents' capabilities. However, each agents' capabilities remains constant. In our work, each agents' `capability' can be thought of as the transportation requests it possesses, which constantly changes between episodes. Thus, our agents must be able to generalise across differing agent capabilities.

\section{Background}
\label{sec: background}

\subsection{Collaborative Vehicle Routing}
\label{sec: collaborative_vehicle_routing}

\begin{figure}
  \captionsetup[subfigure]{justification=centering}
  \centering
  \subfloat[Pre-collaboration (total cost: 3.35)]{
    \includegraphics[width=0.33\linewidth]{pre_collab_tours.pdf}
    \label{fig: aa}
  }
  \subfloat[Post-collaboration with grand coalition $\{1, 2, 3 \}$ (total cost: 2.47)]{
    \includegraphics[width=0.32\linewidth]{post_111_tours_no_axis_labels.pdf}
    \label{fig: ab}
  }
  \subfloat[Post-collaboration with coalition $\{1, 2\}$ (Agent 3 is excluded from the coalition, total cost: 2.59)]
    {\includegraphics[width=0.32\linewidth]{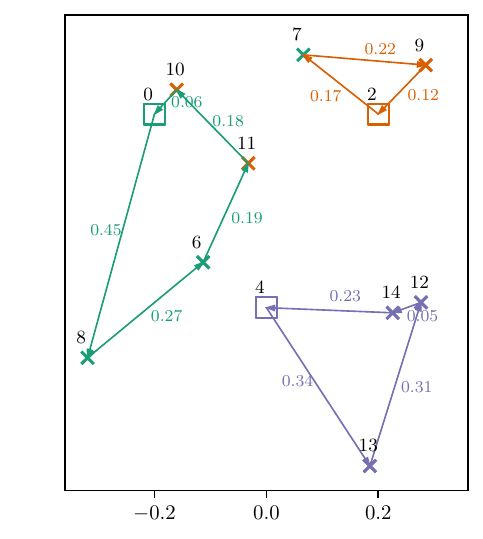}
    \label{fig: ac}
  }
  \caption{Three agents, Agents 1, 2 and 3 are denoted by the colours green, orange and purple respectively. Squares denote depots. Crosses denote customer locations. Node indices (arbitrary) are denoted in black, with costs given in their respective colors. The \emph{collaboration gain} is defined as the difference in social welfare before and after collaboration. \autoref{fig: ab} and \autoref{fig: ac} refer to two possible post-collaboration scenarios with collaboration gains per capita of 0.29 and 0.38 respectively. Thus, it would be rational for the coalition $\{1, 2\}$ to form instead of the grand coalition $\{1, 2, 3\}$.}
  \label{fig: a}
\end{figure}

We denote the set of $n$ agents as $N = \{1, \dots, n\}$. A coalition is a subset of $N$, i.e. $C \subseteq N$. The grand coalition is where all agents are in the coalition, i.e. $C = N$.

\textbf{Pre-collaboration profit and social welfare}: The \emph{pre-collaboration profit} of Agent 1 in \autoref{fig: a} is calculated as follows: the \emph{Revenue} is 3 (1 for each delivery); the \emph{Cost} is 1.42 (sum of the edge distances); thus the \emph{Profit} is 1.58 (Revenue subtract Cost). Similarly, the pre-collaboration profit of Agents 2 and 3 is 2 and 2.07. The \emph{pre-collaboration social welfare} is the sum of the pre-collaboration profits, thus $1.58 + 2 + 2.07 = 5.65$.

\textbf{Post-collaboration ``profit'' and social welfare}: Assuming agents agree to form the grand coalition $C = \{1, 2, 3\}$, the post-collaboration ``profit'' of \mbox{Agent 1} can be calculated as $1 - (0.06 + 0.06) = 0.88$. Note that the post-collaboration ``profit'' for Agent 1 appears to have decreased from 1.58 to 0.88 as a result of collaboration. This will be accounted for when discussing the characteristic function and thus Agent 1 will not lose out when we calculate its reward. For Agents 2 and 3, the post-collaboration ``profit'' is 2.19 and 3.46 respectively. Thus a \emph{post-collaboration social welfare} of $0.88 + 2.19 + 3.46 = 6.53$.

\textbf{Collaboration gain}: The \emph{collaboration gain} is defined as the difference in social welfare before and after collaboration for a given coalition, in this case $6.53 - 5.65 = 0.88$ for the grand coalition. Note that the collaboration gain is always greater than or equal to 0. The \emph{value per capita} is $\frac{0.88}{3} = 0.29$. During the bargaining process, agents are able to choose how to divide this collaboration gain amongst themselves. In the unique case where agents agree to divide the collaboration gain equally, i.e. according to the value per capita, we refer to this as \emph{equal gain sharing}. Note that if only Agents 1 and 2 form a coalition (and exclude Agent 3), then the collaboration gain (assuming equal gain sharing) is divided by 2 instead \--- thus making it rational to object and form the coalition $\{1, 2\}$ (the value per capita of this coalition is 0.38).

\textbf{Characteristic function}: The characteristic function, $v: \mathbf{2}^N \to \mathbb{R}$ calculates for every possible coalition the collaboration gain. Importantly, to fully evaluate the characteristic function would require solving a variant of the Vehicle Routing Problem for every possible coalition which scales $\mathcal{O}(2^n)$.

Following the example in \autoref{fig: a}:

\begin{center}
  \renewcommand{\arraystretch}{1.2}
  \begin{tabular}{l l}
    $v(\{1, 2, 3\}) = 0.88$ & Value per Capita = $\frac{0.88}{3} = 0.29$ \\
    $v(\{1, 2\}) = 0.76$ & Value per Capita = $\frac{0.76}{2} = 0.38$  \\
    $v(\{1, 3\}) = 0.24$ & Value per Capita = $\frac{0.24}{2} = 0.12$  \\
    $v(\{2, 3\}) = 0.01$ & Value per Capita = $\frac{0.01}{2} = 0.005$  \\
  \end{tabular}
\end{center}

It is important to note that the characteristic function is \emph{0-normalised}, \emph{essential} and \emph{super-additive} (see \autoref{sec: coalitional_games} for a formal definition). This guarantees that agents will not lose profits as a result of collaboration. The final \emph{take-home profit} that each agent (or carrier) receives can then be calculated as the sum of the pre-collaboration profit and its respective allocation of the collaboration gain. For Agents 1, 2 and 3, this would equate to $1.58 + \frac{0.88}{3} = 1.87$, $2 + \frac{0.88}{3} = 2.29$ and $2.07 + \frac{0.88}{3} = 2.36$ respectively (assuming equal gain sharing). In reality, carriers will receive this take-home profit (which is always greater than or equal to the pre-collaboration profit) as an incentive to collaborate.

\subsection{Coalitional games}
\label{sec: coalitional_games}

\begin{figure}
    \centering
    \includegraphics[width=0.575\linewidth]{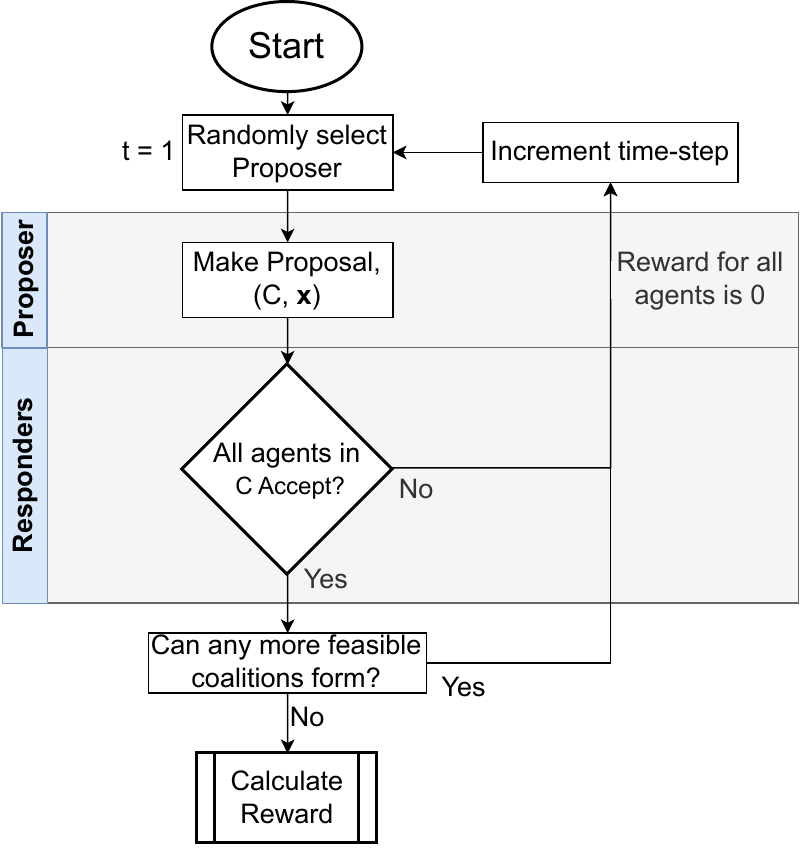}
    \caption{Flowchart of the $n$-player coalitional bargaining game \citep{okada_noncooperative_1996}. Our proposed approach is therefore to obtain a set of intelligent agents that can bargain with each other in a coalitional bargaining game. To achieve a suitable level of agent intelligence, we train our agents using deep multi-agent reinforcement learning.}
    \label{fig: flowchart}
\end{figure}

We consider the $n$-player coalitional game, also called a cooperative game, with a set of agents $N = \{1, \dots, n\}$. A \emph{coalition} is defined as a subset of N, i.e. $C \subseteq N$. The set of all coalitions is denoted $\Sigma$. The \emph{grand coalition} is where the coalition consists of all agents in N, i.e. $C = N$. A \emph{singleton coalition} is where the coalition consists of only one agent, i.e. $|C| = 1$. A \emph{coalition structure} $CS = \{C^1, \dots, C^k\}$ is a partition of $N$ into mutually disjoint coalitions, $C^1 \cup \dots \cup C^k = N$ and $C^i \cap C^j = \varnothing, \forall i \neq j$.

A (transferable utility) coalitional game is a pair $G = \langle N, v \rangle$. The \emph{characteristic function} $v: \mathbf{2}^N \to \mathbb{R}_{\geq0}$ represents the \emph{value} (or collaboration gain in our setting) that a given coalition $C$ receives. Like \citet{okada_noncooperative_1996}, we assume that the characteristic function is \emph{0-normalised}, \emph{essential} and \emph{super-additive}. The characteristic function is \emph{0-normalised} if the value of all singleton coalitions is 0, i.e. $v(\{i\}) = 0, \forall i \in N$. It is \emph{essential} if the value of the grand coalition is strictly positive, $v(N) > 0$. It is \emph{super-additive} if $v(C \cup D) \geq v(C) + v(D)$ for all coalition pairs $C, D \in \Sigma$ where $C \cap D = \varnothing$.

The payoff vector $\mathbf{x}^C = (x_i^C)_{i \in C} $ denotes the pay-off for player $i$ in the coalition $C$. The payoff vector is \emph{feasible} if $\sum_{i \in C} x_i^C \leq v(C)$. The set of all feasible payoff vectors for a given coalition C is $X^C$, and $X_+^C$ when all the elements of $X^C$ is non-negative.

\subsection{Coalitional bargaining}
\label{sec: coalitional_bargaining}

The purpose of this work is to find a partition of the $N$ carriers with an associated payoff vector, i.e. $(CS, \mathbf{x})$, which all self-interested, rational carriers agree to. Notice how this does not imply any sequential decision making. However, it was found that certain cooperative solution concepts can be retrieved as the outcome of non-cooperative, extensive form games such as coalitional bargaining \citep{nash_two-person_1953}. Therefore, this necessitates sequential decision making in our problem where we propose to obtain intelligent agents through the use of MARL.

\citet{okada_noncooperative_1996} presents the $n$-player, random proposers, alternating offers coalitional bargaining game which we adopt. At every time-step  $t = 1, 2, \dots$ an agent from N is selected uniformly at random to be the \emph{proposer}.  The proposer, player $i$, has two actions \--- the proposed coalition and proposed pay-off vector. The proposed coalition $C$ must contain player $i$ and the value of the coalition $v(C)$ must be greater than 0. Due to the characteristic function being 0-normalised this implies $|C| \geq 2$. The payoff vector $\mathbf{x}^C$ must be in the set of all feasible, non-negative payoff vectors $X_+^C$. After player $i$ has proposed, the remaining players called the \emph{responders} are uniformly at random selected sequentially to either accept or reject the proposal. If all agents in the proposed coalition $C$ accepts, then those agents form a coalition with the agreed upon proposal. The remaining players outside of $C$ continue negotiating from the next time-step. If any responder in $C$ rejects the proposal, then all players receive an immediate reward of zero and negotiations go on to the next round of bargaining. Then, a new proposer is selected uniformly at random and the time-step incremented by 1. This continues until either agreement is reached, or the maximum time step is reached. When a proposal $(C, x^C)$ is agreed upon at time $t$, every agent $i$ in $C$ receives a reward of $\gamma^{t-1} x_i^C$, where $\gamma \in [0, 1]$ is the discount factor. The discount factor decreases the reward received as time passes. This encourages agents to reach agreement within the first time-step in the three-player setting as shown in \citet{okada_noncooperative_1996}. The discount factor in this setting is analogous to the \emph{patience} of an agent, or the urgency of the delivery decision. Any agent who is not in a coalition at the end of this process is assumed to have a reward of zero. In the three-player setting, note that if one proposal is accepted, then no more feasible coalitions can form; thus, this denotes the end of the bargaining process as seen in \autoref{fig: flowchart}.

\begin{table}
    \tbl{Notation Table}
    {
        {
            \begin{tabular}{c l}
            \toprule
                
            \bf{Symbol} & \bf{Definition} \\ \midrule
            
            \multicolumn{2}{l}{\textbf{Coalitional Bargaining Game:}} \\
    
            $n$ & Number of Agents \\
    
            $N$ & Set of all $n$ Agents (i.e., grand coalition) \\
    
            $i$ & Agent index \\
            
            $C$ & A Coalition \\
    
            $\Sigma$ & Set of all Coalitions \\
    
            $CS$ & A Coalition Structure \\
    
            $\varnothing$ & Empty set \\
    
            $G$ & A coalitional game \\
    
            $v(\cdot)$ & Characteristic function \\
    
            $v(C)$ & Value of the coalition $C$, or the collaboration gain of the coalition $C$ in the collaborative vehicle routing setting.\\
    
            $\mathbf{x}^C$ & Payoff vector for a given coalition $C$ \\
    
            $X^C$ & Set of all feasible payoff vectors for a given coalition $C$ \\
    
            $X_+^C$ & Set of all feasible, non-negative payoff vectors for a given coalition $C$ \\
            
            \\
    
            \multicolumn{2}{l}{\textbf{(Multi-agent) Reinforcement Learning:}} \\
    
            $\gamma$ & Discount factor \\
    
            $\mathcal{M}$ & A Markov decision process (MDP) \\
    
            $\mathcal{S}$ & Set of states \\
    
            $s_0$ & Initial state of an episode \\
    
            $\mathcal{A}$ & Set of (joint) actions \\ 
    
            $\mathcal{T}$ & Transition probability distribution \\
    
            $\rho_0$ & Distribution of the initial state, $s_0$ \\
    
            $a$ & An action \\
    
            $t$ & Time-step index \\
    
            $G_t$ & Return following time $t$ \\
    
            $T$ & Maximum time-step (or the horizon length) \\
    
            $\pi$ & Agent's policy \\
    
            $V_\pi(s)$ & State-value function of a state $s$ following a policy $\pi$ \\
    
            $\hat{V}(s, \theta)$ & Policy's (parameterised by $\theta$) estimate of the state-value function given the state $s$. \\
    
            $\hat{Q}(s, a, \theta)$ & Policy's (parameterised by $\theta$) estimate of the action-value function given the state $s$ and taking the action $a$. \\
    
            $Q_\pi (s, a)$ & Action-value function of a state $s$ taking the action $a$ following a policy $\pi$ \\
    
            $\mathcal{R}$ & Set of all possible rewards \\
    
            $R_{i, t}$ & Reward at time $t$ for agent $i$ \\
    
            $\theta_i$ & Agent i's policy parameters, usually the parameters of a neural network \\
    
            $J( \theta )$ & Performance measure for the policy $\pi_\theta$ \\
    
            $\nabla J( \theta )$ & Column vector of partial derivatives of $\pi(a | s, \theta)$ with respect to $\theta$ \\
    
            $\hat{g}$ & Estimate of the policy gradient \\
    
            $M$ & Number of episodes played in parallel \\
    
            $\alpha$ & Learning rate for stochastic gradient descent \\
    
            $b(s)$ & A baseline function for policy gradient methods \\
    
            $r_t(\theta)$ & PPO's probability ratio between the new policy (after gradient updates) and the old policy (before gradient updates) \\
    
            $\varepsilon$ & Threshold to clip the probability ratio in PPO \\
    
            $\mathcal{H}$ & Entropy bonus \\
    
            \\
            
            \multicolumn{2}{l}{\textbf{Collaborative Vehicle Routing:}} \\
    
            $\mathbf{D}$ & Deliveries matrix \\
    
            $x$ & x-coordinate of the location \\
    
            $y$ & y-coordinate of the location \\
    
            $o$ & Agent index who owns the location \\
    
            $d$ & Binary variable denoting whether the location is a depot or a customer \\
    
            $\mathbf{c}$ & Multi-hot encoded vector denoting which agents are in the proposed coalition \\
    
            $\mathbf{x}$ & Proposed pay-off vector \\
    
            $\mathbf{r}$ & Responses of the agents to the given proposal \\
    
            $p$ & Agent index who was selected to propose in the current round of bargaining \\
    
            $a$ & Binary variable denoting whether the current agent is proposing or responding \\
    
            Dir$(\boldsymbol{\alpha})$ & Dirichlet distribution with concentration parameters $\boldsymbol{\alpha}$ \\
    
            \bottomrule
            \end{tabular}
        }
    }
    \label{table_2}
\end{table}

\section{Methodology}
\label{sec: methodology}

In summary, analytically calculating cooperative game theory solution concepts is intractable for settings with more than 6 carriers \citep{cruijssen_cross-chain_2020}. Instead, we can recover these cooperative solution concepts through non-cooperative, extensive form games such as coalitional bargaining \citep{serrano_fifty_2004}. However, coalitional bargaining requires intelligent, rational agents and it is difficult to manually craft rule-based agents for collaborative routing due to its exponential and NP-hard nature. Instead, we propose to develop intelligent, rational agents through having agents learn through trial-and-error, learning to collaborate in the presence of multiple other self-interested, rational agents (i.e., multi-agent reinforcement learning). A holistic diagram to depict the whole pipeline can be found in \autoref{sec: holistic_diagram}. The remainder of this section focuses on the reinforcement learning algorithm employed. Pseudo-code of the pipeline can be found in \autoref{sec: pseudo_code}.

\subsection{Single Agent Reinforcement Learning}
Reinforcement Learning (RL) is a subfield of machine learning. Here, the field studies an agent learning what \emph{actions} to take for a given \emph{state} in order to maximise a numerical \emph{reward}. In supervised learning, the ground truth target labels are provided. In RL, we are not told the ``correct'' actions to take that will maximise (expected) cumulative reward. Instead, the agent must learn through trial-and-error. This leads to an exploration-exploitation dilemma. Should the agent try new actions (explore) in the hope that there is a better sequence of actions that leads to an even higher expected reward? Or, should the agent stick with its current best-known actions (exploit) since the agent believes it is unlikely there will be a better sequence of actions with higher expected reward? \citep{sutton_reinforcement_2018}. The agent selects actions according to its \emph{policy} based on the current state. The action is sent to the \emph{environment} which calculates the reward and next state which is then returned to the agent. Through the learning process, we aim to obtain a policy that maximises the expected cumulative reward.

In our setting of collaborative vehicle routing, the environment is the coalitional bargaining game as described in \autoref{sec: coalitional_bargaining}. Each carrier is represented as an individual agent. The state is the locations of depots and customers, as well as auxiliary features to describe the current state of the coalitional bargaining process \--- see \autoref{sec: state_space} for further details. There are three actions that an agent can take depending on if it is proposing or responding. When proposing, the agent must decide (a) which other carriers should the agent propose to partner with, and (b) how much should each carrier in the proposal be paid. When responding, the agent must decide (c) if they accept or reject the proposal. The reward is the collaboration gain the agent is allocated as a result of the coalitional bargaining process. Throughout the training process, we train our agents' policies (or neural network) to maximise expected cumulative reward. See \autoref{sec: methodology} for a formal definition of states, actions and rewards in our setting.

We can formalise the problem using Markov decision processes (MDPs) \citep{puterman_markov_1994}. Formally, a finite-horizon, discounted Markov decision process $\mathcal{M}$ can be defined by the tuple $\mathcal{M} = \langle \mathcal{S}, \mathcal{A}, P, r, \rho_0, \gamma \rangle$ where $\mathcal{S}$ is the set of states, $\mathcal{A}$ is the set of actions, $\mathcal{T}: \mathcal{S} \times \mathcal{A} \to \mathcal{S}$ is the transition probability distribution, $\mathcal{R}: \mathcal{S} \times \mathcal{A} \times \mathcal{S} \to \mathbb{R}$ is the reward function, $\rho_0 : \mathcal{S} \to \mathbb{R}$ is the distribution of the initial state $s_0$, and $\gamma \in [0, 1]$ is the discount factor.

An episode begins by first sampling an initial state $s_0$ from $\rho_0$. A \emph{trajectory} $(s_0, a_0, s_1, a_1, \dots)$ is generated by sampling actions from the agent's policy $a_t \sim \pi(a_t \, | \, s_t)$. The next states are obtained by sampling the transition dynamics function $s_{t+1} \sim \mathcal{T}(s_{t+1} \, | \, s_t, a_t)$ until reaching a terminal state. At each time step, a reward $R_t \sim \mathcal{R}(s_t, a_t, s_{t+1})$ is received. At timestep $t$, the discounted return, $G_t$, is defined as:

\begin{equation}
    G_t \doteq R_{t+1} + \gamma R_{t+2} + \gamma^2 R_{t+3} + \dots + \gamma^{T} R_{T+1}= \sum_{k=0}^T \gamma^k R_{t+k+1}
\end{equation}

where $T$ is the maximum time-step and $\gamma \in [0, 1]$ is the discount factor. As $\gamma$ approaches 1, the agent will take into account rewards received far into the future. However, as $\gamma$ approaches 0, the agent will only account for the immediate reward $R_{t+1}$, and the agent is often said to be \emph{myopic}.

The \emph{state-value function} of a state $s$ under a policy $\pi$ is denoted by $V_\pi(s)$. This is the expected return when the agent starts in $s$ and continues following its policy $\pi$. Formally:

\begin{equation}
    V_\pi(s) \doteq \mathbb{E}_\pi \left[G_t \, | \, S_t = s \right] = \mathbb{E}_\pi \left[ \sum_{k=0}^T \gamma^k R_{t+k+1} \, | \, S_t = s  \right], \qquad \forall s \in \mathcal{S}
\end{equation}

A similar notion is the \emph{action-value function} which is denoted by $Q_\pi (s, a)$. This is the expected return when the agent starts from $s$, but also takes the action $a$, and follows its policy $\pi$ afterwards. Formally:

\begin{equation}
    Q_\pi(s, a) \doteq \mathbb{E}_\pi \left[G_t \, | \, S_t = s, A_t = a \right] = \mathbb{E}_\pi \left[ \sum_{k=0}^T \gamma^k R_{t+k+1} \, | \, S_t = s, A_t = a  \right]
\end{equation}

\subsection{Multi-agent reinforcement learning}
A \emph{stochastic game} generalises MDPs to involve multiple agents. This can be defined as a tuple $\langle N, S, A, \mathcal{T}, \mathcal{R}, \gamma, \rangle$ where:

\begin{itemize}
  \item $N$ denotes the set of $n$ agents
  \item $S$ denotes the set of states including the initial state $s_0$
  \item $A = A_i \times \dots \times A_n = \{(a_1, \dots a_n)\, |\, a_i \in A_i\, \text{for every}\,  i \in \{1, \dots, n\} \}$ denotes the set of joint actions, where $A_i$ is player $i$'s set of actions and $\times$ denotes the Cartesian product.
  \item $\mathcal{T}: S \times A \to S$ denotes the transition dynamics
  \item $\mathcal{R}: S \times A \times S \ \times\ N \to \mathbb{R}$ denotes the reward function
  \item $\gamma$ denotes the discount factor
\end{itemize}

For every time-step $t$, an agent $i \in N$ receives an observation of the global state $s$ and outputs an action $a_{i, t}$ sampled from its \emph{policy} $\pi_{i}(a_{i, t} \mid s_t)$. We update the state $s_t$ to include agent $i$'s action before sending this new state to agent $j \in N, j \neq i$. Note that the time-step is not yet incremented. We continue this process until all agents in $N$ have submitted their actions to the environment. This yields the joint action $\mathbf{a} = (a_1, \dots a_n)$. We calculate the reward $R_{i, t} \sim \mathcal{R}(s_{t}, \mathbf{a}, s_{t + 1}, i)$. We consider the sparse reward setting, i.e., all rewards are zero until the episode terminates. Upon termination, we calculate the reward for agent $i$ depending on if agent $i$ successfully joined a coalition or not. When a proposal $(C, x^C)$ is agreed upon at time $t$, every agent in $C$ receives a reward of $\gamma^{t-1} x_i^C v(C)$. Else, if the agent is not in a coalition $C$, it is assumed to receive a reward of zero. The return $G_i$ is discounted by a factor $\gamma \in [0, 1]$, given by $G_i = \sum_{t = 1}^{T} \gamma^{t-1} r_{i, t}$.

Agent $i$'s objective is to find a policy $\pi_{\theta_i}$ which maximises its expected discounted sum of rewards $\mathbb{E}[\sum_{t=1}^T \gamma^{t-1}R_{i, t}]$. It is important to note that this maximisation assumes all opponents' policies $\pi_{\theta_j} \; \forall j \neq i$ to be fixed. Thus, one of the key challenges in MARL is the non-stationarity present due to multiple concurrently learning agents.

In our setting, we assume perfect information and thus agents have full access to the global state. We make this assumption as the aim of our paper is to provide the theoretical grounding between collaborative vehicle routing, coalitional bargaining, and multi-agent reinforcement learning. The imperfect information setting is also a promising research direction, e.g., to investigate the value of information. Future work could study the applicability of \emph{decentralised partially observable} Markov decision processes (dec-POMDPs) \citep{oliehoek_concise_2016} to imperfect information settings in collaborative vehicle routing.

A challenge in reinforcement learning is handling the curses (plural) of dimensionality \citep{powell_reinforcement_2022}. With ``tabular'' methods, the policy is represented by a lookup table. One curse is that the size of the state space grows exponentially with the number of dimensions (even if the state space is discrete). In our setting, our state space is continuous thus further exacerbating the challenge. As a result, we must resort to \emph{function approximation} methods \citep{sutton_policy_2000}. Instead, we aim to replace the lookup table with a parameterised model, with parameters $\theta \in \mathbb{R}^d$ to map from states to actions. Thus, we can write the policy for agent $i$ as $\pi_{\theta_i}(a_{i, t} \, | \, s_t)$ instead. Respectively, the state-value function and action-value function can also be re-written $\hat{V}(s, \mathbf{\theta}) \approx V_\pi(s)$ and $\hat{Q}(s, a, \mathbf{\theta}) \approx Q_\pi (s, a)$. Importantly, the dimensionality $d$ of the model is typically much less than the number of states. Changing one parameter will effect the estimated value of many other states. Thus, if we can generalise across states, this could greatly accelerate learning. Note that any parameterised model can be used: a linear function, multi-layer perceptron, decision trees etc. Historically, linear functions were favoured due to favourable convergence guarantees. However, deep neural networks have demonstrated significant success due to their high capacity and generalisability \citep{sutton_reinforcement_2018, vinyals_grandmaster_2019, mnih_human-level_2015, openai_dota_2019}. Thus, we also opt for deep neural networks as well.

\emph{Policy gradient}-based approaches are a common way to learn a parameterised policy $\pi_{\theta_i}$ which maximises an agent's expected discounted return. It is also performant, for example, it achieved great success in playing Dota 2 \citep{openai_dota_2019} amongst others. Typically, a scalar performance measure $J(\mathbf{\theta})$ is defined and we maximise their performance using approximate gradient ascent: $\mathbf{\theta}_{t+1} = \mathbf{\theta}_t + \alpha \widehat{\nabla J(\mathbf{\theta}_t})$ where $\widehat{\nabla J(\mathbf{\theta}_t}) \in \mathbb{R}^d$ is a stochastic estimate whose expectation approximates the gradient of $J(\theta_t)$ with respect to $\theta_t$. However, a challenge is that the performance depends on both the policy's action selection and also the distribution of states where these actions are selected. Varying $\theta$ affects both of these distributions and we typically do not know the effect of our policy on the state distribution. The policy gradient theorem \citep{sutton_policy_2000, sutton_reinforcement_2018} shows that we can approximate the gradient of performance with respect to $\theta$ but without requiring the derivative of the state distribution. Formally:

\begin{equation}
    \nabla J(\theta) \propto \sum_s \mu(s) \sum_a Q_\pi(s, a) \nabla \pi(a\, | \, s, \theta)
\end{equation}

The simplest approach is the REINFORCE algorithm \citep{williams_simple_1992}. Here, an agent plays $M$ episodes in parallel until termination and remembers all states, actions and rewards it encountered (or trajectory). Next, it estimates the (undiscounted) policy gradient using:

\begin{equation}
  \hat{g} = \frac{1}{M} \sum_{m=1}^M \left[ \sum_{t=1}^T \hat{A}_t^m \nabla_\theta \log \pi_\theta (a_t^m\, |\, s_t^m)  \right]
\end{equation}

where, for REINFORCE $\hat{A}_t = \sum_{t'=t}^T \gamma^{t'-t} r(s_{t'}^m, a_{t'}^m) $. The agent updates its policy using stochastic gradient descent, i.e., $\theta \leftarrow \theta + \alpha \hat{g}$ where $\alpha$ is the learning rate. The intuition for this policy update is that for each action the agent took for a given state, it will increase or decrease the (log) probability of taking that same action proportional to the discounted return it received during that episode. However, policy gradient methods are notorious for having high variance in the policy gradient. As a result, we employ multiple variance reduction techniques to mitigate this problem, such as $M$ parallel environments.

Another variance reduction technique is to subtract a \emph{baseline}. A baseline $b(s)$ can be any function that may or may not depend on the state $s$. Importantly, it must not vary with the action $a$. We can replace REINFORCE's estimate of $\hat{A}_t$ by using $\hat{A}_t = \left[ \left( \sum_{t'=t}^T \gamma^{t'-t} r(s_{t'}^m, a_{t'}^m) \right) - b(s) \right] $ instead. It can be shown that introducing a baseline does not introduce bias into the policy gradient, but may significantly reduce variance \citep{williams_simple_1992,greensmith_variance_2004,sutton_reinforcement_2018}. An example baseline is the average return an agent received. The term $ \left[ \left( \sum_{t'=t}^T \gamma^{t'-t} r(s_{t'}^m, a_{t'}^m) \right) - b(s) \right] $ can be thought of as how much better than the baseline an agent performed as a result of choosing its action. A common choice of $b(s)$ is to estimate the state-value {$\hat{V}_\pi(s_t^m, \theta) = \mathbb{E}_\pi \left[ \sum_{t'=t}^T \gamma^{t'-t} r(s_{t'}^m, a_{t'}^m) \, | \, S_t = s  \right]$. Selecting a good baseline is crucial. We discuss our proposed baseline functions in \autoref{sec: baseline}.

In REINFORCE, typically only one gradient update is used per batch of trajectories. As a result, REINFORCE is typically said to be sample inefficient \--- it requires a lot of episodes to train a performant policy. In addition, REINFORCE can be unstable during training, and sometimes performance collapse may occur as a result of the data distribution changing too drastically.

Proximal Policy Optimisation (PPO) \citep{schulman_proximal_2017} aims to improve the sample efficiency by performing multiple gradient updates to maximise the use of each gathered data point. However, this risks changing the data distribution too drastically and thus risks performance collapse. To rectify this, the intuition behind PPO is to constrain the policy from deviating too greatly. Let the current policy (before any gradient updates) be denoted $\pi_{\theta_{old}}(a_t |s_t)$. After one round of gradient updates, this would yield new policy parameters, denoted $\pi_\theta (a_t | s_t)$. PPO constrains that the probability ratio, $r_t(\theta) = \frac{\pi_\theta (a_t | s_t)}{\pi_{\theta_{old}} (a_t | s_t)}$, of taking action $a_t$ for the same state $s_t$ under the old policy vs new policy to be no more than a certain percentage $\varepsilon$. This should prevent the risk of policy collapse if $\varepsilon$ is chosen carefully. Moreover, PPO is then able to perform more gradient updates on the same data points, thus greatly improving its sample efficiency. In addition, it is also more stable during training and is less sensitive to chosen hyperparameters. As a result, PPO has been applied to wide range of domains, most notably in OpenAI Five (bots to play Dota 2) \citep{openai_dota_2019} and also in ChatGPT \citep{openai_chatgpt_2022}.

PPO  adjusts the neural network parameters $\theta$ to increase or decrease the probability ratio $r_t(\theta)$ proportional to the advantage the agent received $\hat{A}_t$. PPO enforces the $\varepsilon$ threshold by clipping the probability ratio, $r_t(\theta)$, to remain within $\pm \, \varepsilon$. We can further encourage exploration by adding an entropy bonus. Thus, the PPO policy gradient can be estimated as follows:

\begin{multline}
    \hat{g} \approx \frac{1}{M} \sum_{m=1}^M \sum_{t=1}^T \nabla_\theta \Bigl[ \min (r_t(\theta)\hat{A}_t, \, \text{clip}(r_t(\theta), 1 - \varepsilon, 1 + \varepsilon) \, \hat{A}_t) + \beta \mathcal{H}[\pi_\theta](s_t) \Bigr] 
\end{multline}

where $\hat{A}_t$ is the baseline, $\beta$ is the entropy regularisation coefficient and $\mathcal{H}$ the entropy bonus. An entropy bonus encourages agents to explore rather than exploit. It is important to note that when the advantage is positive, we clip $r_t(\theta)$ \emph{only} if it is greater than $1 + \varepsilon$. If the advantage is negative, we clip $r_t(\theta)$ \emph{only} if it is less than $1 - \varepsilon$ (see Figure 1 of \citep{schulman_proximal_2017} for further details). The \texttt{clip} function is a function that clips the first argument by the lower and upper bounds denoted by the second and third arguments respectively.

As a result, PPO has been widely used in a range of applications, most notably in OpenAI Five (for Dota 2) and in ChatGPT \citep{openai_dota_2019, openai_chatgpt_2022}.

\subsection{State space}
\label{sec: state_space}

The agents receive a variety of inputs from the environment as seen in \autoref{fig: actor_nn}. Let the state at time $t$ be denoted by $s_t \in S$ which can be represented by the tuple $\langle \mathbf{D}, \mathbf{c}, \mathbf{x}, \mathbf{r}, t, p, a \rangle$. The \emph{deliveries} matrix $\mathbf{D} \in \mathbb{R}^{12 \times 4}$ describes the features of each of the three depots and nine customers, yielding twelve rows where we refer to each row as a \emph{location}. A location can be represented by the tuple $\langle x, y, o, d \rangle$ where $x \in \mathbb{R}$ is the x-coordinate; $y \in \mathbb{R}$ is the y-coordinate; $o \in \mathbb{N}$ denotes the agent who owns the location; and $d \in \{0, 1\}$ denotes whether the location is a depot or a customer. For instance, to represent Agent 2's depot located at $\langle x = 0.2, y = 0.173 \rangle$, its corresponding row in $\mathbf{D}$ would be represented as $\langle 0.2, 0.173, 2, 1 \rangle$ and the remaining rows in $\mathbf{D}$ would be comprised of similar entries for the remaining depots and customers, yielding a shape of $12 \times 4$. The vector $\mathbf{c} \in \{0, 1\}^{|N|}$ denotes which agents were selected to be in the proposed coalition. The vector $\mathbf{x} \in \mathbb{R}^{|N|}$ denotes the proposed pay-off vector, the vector $\mathbf{r} \in \{0, 1\}^{|N|}$ denotes the responses of the agents. The vectors $\mathbf{c}$, $\mathbf{x}$ and $\mathbf{r}$ are initialised to zero if no agent has taken an action in the current round of bargaining. The scalar $t \in \mathbb{N}_0$ denotes the current round of bargaining, $p \in \mathbb{N}$ denotes which agent was selected to propose in the current round of bargaining, and $a \in \{0, 1\}$ denotes whether the current agent is proposing or responding.

\begin{figure}[t]
    \centering
    \includegraphics[width=0.95\linewidth]{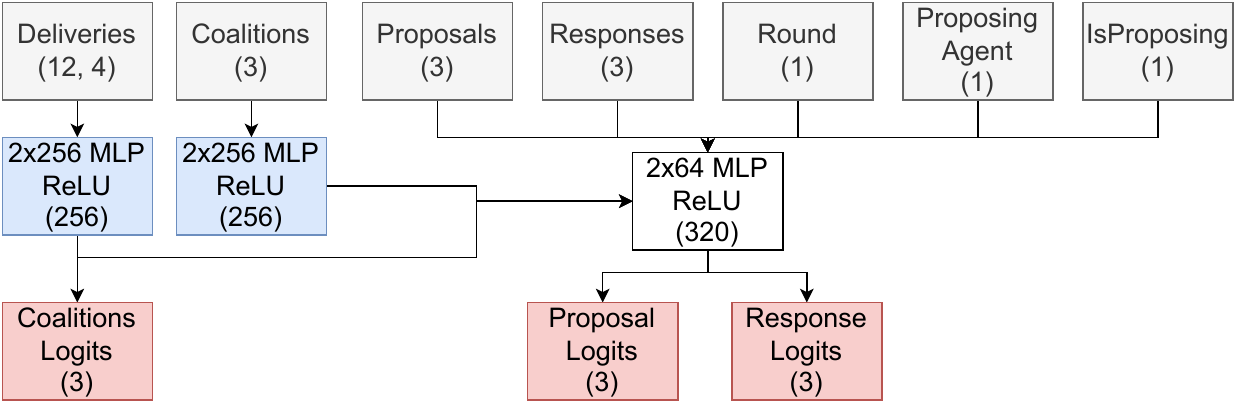}
    \caption{Actors' neural network design. Grey boxes denote state inputs.
    Blue boxes denote MLP parameters which come from supervised pre-training (see \autoref{sec: supervised}). Note that the linear layer to produce coalition logits is learnt and not pre-trained. White boxes denote learnt parameters. Red boxes denote actions. Numbers in brackets denote the output shapes (ignoring batch size as it's shared by all).}
    \label{fig: actor_nn}
\end{figure}

\subsection{Action space}
\label{sec: action_space}

The agents have three action heads: \emph{coalitions}, \emph{proposals} and \emph{response}.

The \emph{coalitions} action is denoted by $\mathbf{c} \in \{0, 1\}^{|N|}$ where $|N|$ is the total number of agents, in this case, 3. Note that in game theory, typically agents propose a coalition of size $|C|$ instead of $|N|$. However, it is beneficial to output coalitions in this manner as it keeps the output size constant. The coalitions action denotes whether the respective agent index is part of the coalition $C$. Note that this game assumes that player $i$ is in the coalition $\mathbf{c}$, i.e., $c_i = 1$. The \emph{deliveries} matrix, $\mathbf{D}$, is fed through two dense layers with 256 hidden neurons.} These parameters come from a supervised pre-training step (see \autoref{sec: supervised}). The output is fed through a linear layer with $|N|$ outputs. These outputs are passed into $|N|$ independent Bernoulli distributions to determine the probability that a given agent is in the coalition $C$. A Bernoulli distribution is chosen as the number of outputs required scales linearly with the number of agents. Alternatively, this action can be output auto-regressively, but would be more computationally expensive. It may also be useful to introduce correlation in the agents' actions via more expressive probability distributions which may speed up learning.

The \emph{proposals} action is denoted by $\mathbf{x} \in \mathbb{R}^{|N|}$ where $\sum_i{x_i} = 1, \ x_i \in [0, 1]$. This vector denotes how much of the collaboration gain is assigned to each respective agent (as a percentage). Note that in game theory, the definition of a feasible pay-off vector is $\sum_{i \in C} x_i^C \leq v(C)$. However, agents will never know the value of $v(C)$ a priori (although it can implicitly reason about it). Thus, to practically implement our neural network, we output a vector that is interpreted as percentages as opposed to absolute values. These percentages are then multiplied by the value of a coalition $v(C)$ to obtain a feasible pay-off vector.

Note that this is a continuous action space, as opposed to the other actions which are discrete. To parameterise the proposals action head, we use the Dirichlet distribution which is a multivariate generalisation of the Beta distribution. The neural network will output three logits $\mathbf{\alpha}$ which are used as the concentration parameters of the Dirichlet distribution Dir$(\mathbf{\alpha})$. The Dirichlet distribution has support over the probability simplex $ S_K = \{ \mathbf{\theta}: 0 \leq \theta_{k} \leq 1, \sum_{k=1}^{K} \theta_{k} = 1 \} $ \citep{murphy_probabilistic_2021}. Intuitively, agents will propose an equal gain share with high probability if the inputs to the Dirichlet are large and equal. Agents will make proposals uniformly at random within the probability simplex if the inputs to the Dirichlet are small and equal, but greater than 1. If Agent 1 wanted to collaborate with Agent 2 but not 3, the input to the Dirichlet could be $\langle 10 000, 10 000, 1.001 \rangle$. This would result in approximately a 50/50 split between Agents 1 and 2 with high probability.

The Dirichlet distribution is appealing due to two reasons. Firstly, the proposals vector requires that it sums to 1 which matches the form of the Dirichlet distribution. Secondly, the Dirichlet distribution has finite support. In continuous action spaces, a Gaussian distribution is typically used which has infinite support and can lead to bias \citep{chou_improving_2017}. \citet{chou_improving_2017} overcomes this issue by using a Beta distribution instead as it has finite support and find that their agents learn more efficiently.

To calculate the proposals, the state inputs are passed through a variety of dense layers (see \autoref{fig: actor_nn}) to produce an embedding. A linear layer with 3 output neurons is applied to the embedding. As in \citet{chou_improving_2017} we add 1.001 to the output logits to ensure the resultant Dirichlet distribution remains unimodal. As a result, during evaluation the agents can fully exploit by proposing the mode of the distribution, instead of having to sample from the Dirichlet which may involve exploration. The output logits are then masked by the \emph{coalitions} vector, i.e. if a player $i$ is not in the coalition $S$, its corresponding output logit will be 1.001. Finally, to calculate the pay-off vector, we sample from the Dirichlet distribution with the masked output logits.

The \emph{response} action $r \in \{0, 1\}$ denotes whether an agent accepts or rejects a given proposal. It takes the resultant embedding followed by a single linear layer with one output neuron. The output is then fed through a Bernoulli distribution.

Whilst we have chosen to use Bernoulli and Dirichlet distributions to parameterise the three action spaces, it may be beneficial to experiment with more expressive probability distributions or e.g. output actions auto-regressively. This may speed up learning and would be an interesting line of future research.

\subsection{Reward function}

Our reward function is sparse, i.e., at timestep $t$ the agents will always receive an immediate reward $R_t$ of zero until the coalitional bargaining game terminates. Upon termination, we calculate a reward for each agent.

If agent $i$ successfully joins a coalition $C$ by having all agents in $C$ accept the proposal, then it receives a reward of $r_{i, t} = v(C) \cdot x_i$ where $v(C)$ is the collaboration gain obtained by coalition $C$, and $x_i$ is the $i$th element of the pay-off vector $\mathbf{x}$. For clarity, if agent $i$ is the proposer and has its proposal rejected by the responder agents, it will receive an immediate reward of zero. However, there is potential for agent $i$ to obtain more than zero immediate reward in future rounds of bargaining and thus the discounted return can still be greater than zero.

Else, if agent $i$ does not successfully join a coalition $C$ by the end of the episode, then it will receive a terminal reward of zero.

\subsection{Transfer learning}

A key challenge with policy gradient approaches is its sample inefficiency, even in single agent settings. This is further exacerbated due to the non-stationary learning dynamics imposed by having multiple agents learn concurrently. In typical RL settings, agents learn ``tabula rasa'', i.e., without any prior knowledge. Whilst this is mathematically elegant, learning tasks tabula rasa for problems with high complexity, such as in real-world, multi-agent settings, is rare \citep{agarwal_reincarnating_2022}. Instead, it may be preferable to pre-train on some offline dataset in order to learn a good feature extractor. For example, \citep{silver_mastering_2016,vinyals_grandmaster_2019} pre-train their networks on human gameplay data in a supervised learning setting before using RL. This idea of transfer learning, or recently, \emph{reincarnating RL} \citep{agarwal_reincarnating_2022} is well accepted in the RL literature and the reader is referred to \citep{taylor_transfer_2009,agarwal_reincarnating_2022} for a thorough review. Furthermore, transfer learning is well accepted in supervised learning, especially in the computer vision and natural language processing domains leading to the likes of ChatGPT \citep{openai_chatgpt_2022}. In our case, the pre-training process aids in efficiently initializing the agents' policies and facilitates faster convergence in the MARL framework.

\label{sec: supervised}

\begin{figure}
  \centering
  \includegraphics[width=0.25\linewidth]{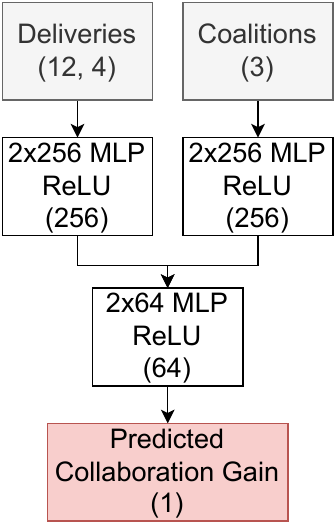}  
  \caption{Pre-trained neural network design. Grey boxes denote state inputs. White boxes denote learnt parameters. Red box denotes the output, which predicts the collaboration gain for this given state and coalition}. Numbers in brackets denote the output shapes (ignoring batch size as it's shared by all).
  \label{fig: supervised}
\end{figure}

We therefore pre-train our agents to learn a good feature extractor in a supervised learning fashion. We hypothesise that a good feature extractor should be able to predict whether a given coalition for a given state is productive or not. As a result, we create a dataset of one million instances and randomly select a feasible coalition per instance and calculate the social welfare obtained. Next, we train a neural network to predict the social welfare for a given state and coalition. We optimise the neural network to minimise the mean squared error. We split the dataset using an 80/20 train/test split. The neural network design can be seen in \autoref{fig: supervised}. We experimented with different neural network architectures but found this architecture performed best. Whilst this is not the exact task agents must perform in the collaborative routing scenario, the intuition is that the neural network should still learn useful patterns which are transferable to the full collaborative routing problem.

\subsection{Policy gradient baselines}
\label{sec: baseline}

\begin{figure}[t]
    \centering
    \subfloat[Neural Network design of the \emph{coalitions} and \emph{proposals} baseline.]
    {
        \resizebox*{0.85\linewidth}{!}
        {
        \includegraphics{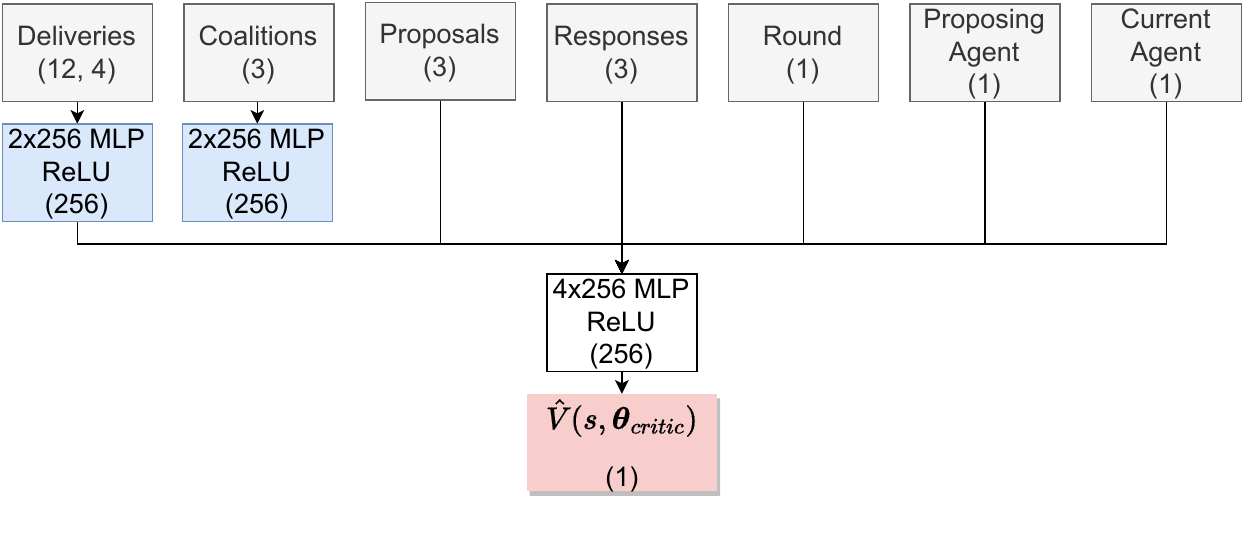}
        \label{fig: proposals_baseline}
        }
    }
    \hspace{5pt}
    \subfloat[Neural network design of the \emph{responses} baseline.]
    {
        \resizebox*{0.85\linewidth}{!}
        {
        \includegraphics{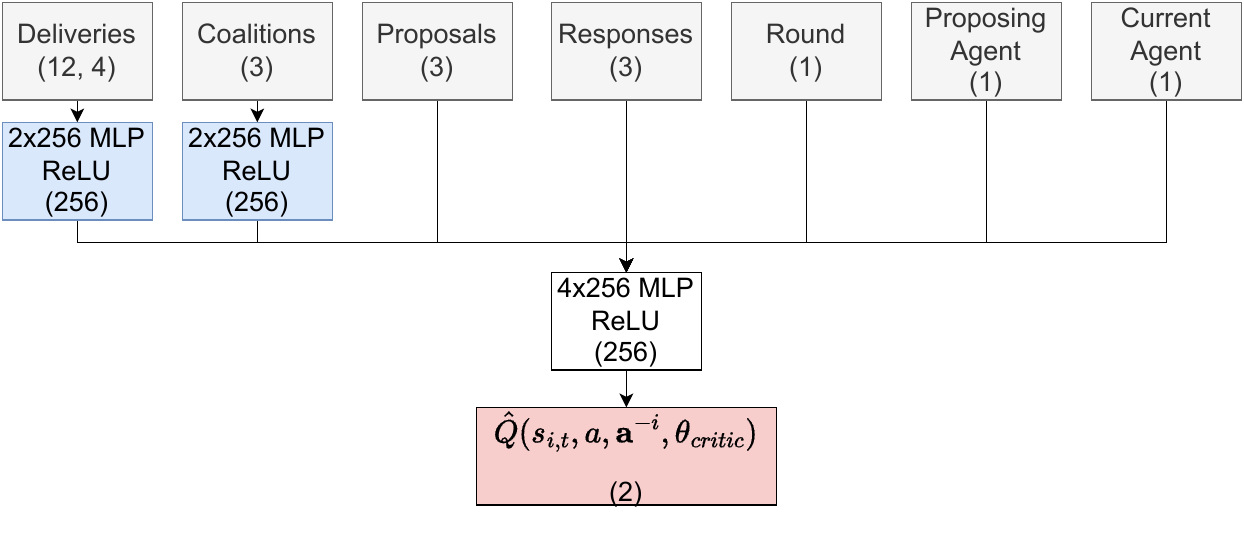}
        \label{fig: responses_baseline}
        }
    }
    \caption{Grey boxes denote state inputs. Blue boxes denote MLP parameters which come from supervised pre-training (see \autoref{sec: supervised}). White boxes denote learnt parameters. Red boxes denote outputs for the baseline. Numbers in brackets denote the output shapes (ignoring batch size as it's shared by all).}
    \label{fig: baselines}
\end{figure}

As discussed in \autoref{sec: methodology}, a useful baseline helps reduce the variance in policy gradient methods. We use two types of baselines: one for the response action (when agents are responding); and one shared for both the coalitions and proposals actions (when agents are proposing). The neural network architectures for the baselines can be found in \autoref{fig: baselines}.

The response action is discrete and thus we can easily implement Counterfactual Multi-Agent Policy Gradients (COMA) \citep{foerster_counterfactual_2017}. They use the following baseline:

\begin{equation}
  A^i(s, \mathbf{a}) = Q(s, \mathbf{a}, i) - \sum_{a^{'i}} \pi^i (a^{'i} | \tau^{i}) \hat{Q}(s, (\mathbf{a}^{-i}, a^{'i}), i, \phi)
\end{equation}

where $Q_{\pi}(s, \mathbf{a}, i) = \mathbb{E}_\pi \left[ \sum_{t'=t}^T \gamma^{t'-t} r(s_{t'}^m, a_{t'}^m) \, | \, s, \mathbf{a} \right] $ is the discounted return received if all agents take the joint action $\mathbf{a}$ in state $s$. The estimate comes from a function approximator with parameters $\phi$. $a^{'i}$ is the other actions agent $i$ could have taken. $\tau^i$ is the prior trajectory agent $i$ has observed. $\hat{Q}(s, (\mathbf{a}^{-i}, a^{'i}), i, \phi)$ is the \emph{estimated} discounted return agent $i$ would receive if it took a different action $a^{'i}$ whilst keeping the other agents' actions $\mathbf{a}^{-i}$ constant. This is estimated through the use of a neural network with parameters $\phi$.

Intuitively, the COMA baseline can be thought of as how much better agent $i$'s decision to take action $a$ was relative to any other action agent $i$ could have taken, $a^{'i}$. In our case, the question we ask is: if an agent has agreed to a given proposal, could it have done better by rejecting instead, assuming other agents' actions remain the same?

In the discrete setting, it is easy to sum over all other actions agent $i$ could have taken. However, with continuous actions using Dirichlet distributions in the proposals action, this can be difficult. Therefore, we instead estimate the \emph{state-value} which estimates the expected discounted return conditioned on the state $s$. We denote this baseline with $\hat{V}_{\pi}(s, i, \mathbf{w}) = \mathbb{E}_\pi \left[ \sum_{t'=t}^T \gamma^{t'-t} r(s_{t'}^m, a_{t'}^m) \, | \, s \right] $ where $\mathbf{w}$ is the parameters of a function approximator such as a neural network. Thus, our baseline for both the coalitions action and proposals action is given by:

\begin{equation}
    A^i(s, \mathbf{a}) = Q_\pi(s, \mathbf{a}, i) - \hat{V}_\pi(s, i, \mathbf{w})
\end{equation}

Finally, we normalise $A^i(s, \mathbf{a})$ by subtracting the mean and dividing by the standard deviation due to the small magnitude in rewards.

\subsection{Time limits}
\label{sec: time_limits}

It is crucial to deal with time limits properly in this setting. The full coalitional bargaining game presented in \citet{okada_noncooperative_1996} is infinite horizon, i.e., negotiation could go on indefinitely. Clearly, this is impossible to simulate on a finite computer and we must set a maximum number of rounds. Nevertheless, it is still possible to optimise for the infinite horizon, but care must be taken as shown in \citet{pardo_time_2018}. They argue that if an episode terminates only due to reaching the maximum number of rounds, we should \emph{bootstrap} the discounted estimated value of the next state, $\hat{v}_\pi(s')$. Thus, if agents reach agreement \emph{within} the maximum number of rounds, they should receive a reward $r$ as expected. However, if they \emph{exceed} the maximum number of rounds, they should receive a reward of $r + \gamma \hat{v}_\pi(s')$. In our setting with the maximum number of rounds equal to 10, if agents do not reach agreement within 10 rounds, we fictitiously step them into the next state $s'$, at round 11 with proposers selected uniformly at random. If player $i$ is not selected as a proposer, then the selected proposer is asked to propose a coalition and pay-off vector $(S, \mathbf{x})$ in this fictitious round. We then use a critic to estimate the value of this state, $\hat{v}_\pi (s')$.

\subsection{Skill retention}
\label{sec: skill_retention}

\citet{okada_noncooperative_1996} shows that agents should reach agreement with no delay in agreement. Therefore, as agents learn to collaborate better, they will reach agreement sooner, which is beneficial due to the environment's discount factor. However, this may lead to agents forgetting how to play the game at later time-steps. To enable retention of skills at later time-steps, we employ a targeted training design. During training, instead of starting all bargaining games at round 1, we uniformly at random start them between round 1 and the last round of bargaining, $T - 1$. Therefore, agents will always be exposed to a range of bargaining scenarios even if agents are collaborating optimally.

\section{Experiments}
\label{sec: exp}

\begin{figure}
  \centering
  \includegraphics[width=0.45\linewidth]{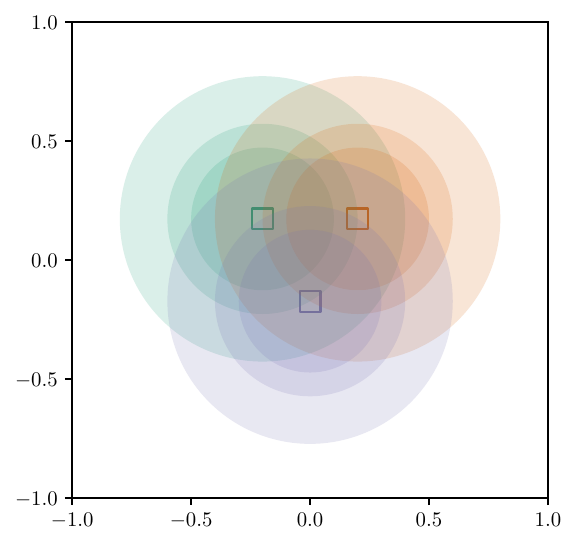}  
  \caption{A plot of the distribution of depot and customer locations. Depots are denoted by squares. Each depot has three distinct service radii which are selected uniformly at random. Customers may be uniformly at random located within any of corresponding depot's service radius.}
  \label{fig: f2}
\end{figure}

\subsection{Problem setting}
We base our problem setting on a modified version of \citep{gansterer_centralized_2018}. We consider an environment with three companies, each represented by an agent. Each agent has one depot and three customers that it must deliver to. The depot $(x, y)$ locations for each agent are held fixed at $\{(-0.2, 0.173), (0.2, 0.173), (0, -0.173)\}$ respectively. The depots' service radius for each instance is selected uniformly at random from the set $\{0.3, 0.4, 0.6\}$. The rationale by \citet{gansterer_centralized_2018} is that, through varying the depots' service radius, this varies the degree of overlap and thus competition (or collaboration opportunity) between carriers. A high degree of overlap using a radius of 0.6 creates high collaboration opportunity between carriers. A low degree of overlap using a radius of 0.3 has low collaboration opportunity between carriers. With a small radius of 0.3, this can analogously be seen as the scenario when depots do not lie in close proximity to each other. The customers locations are then generated uniformly at random with the depot's service radius.

To calculate the pre-collaboration and post-collaboration gains, the shortest paths are calculated exactly using Gurobi \citep{gurobi_optimization_llc_gurobi_2021}. The pre-collaboration shortest paths can be calculated by solving three (un)Capacitated Vehicle Routing Problems (one for each agent). The post-collaboration shortest paths are calculated by solving a single multi-depot vehicle routing problem. Problem formulations for the capacitated VRP and multi-depot VRP can be found in Appendices \ref{app: b} and \ref{app: c} respectively. Capacity is effectively removed by setting the capacity of each vehicle to an arbitrarily large number and the weight of each delivery to 1.

Whilst this problem setting is rather simplistic, this is important as it allows us to evaluate our agents rigorously. To calculate optimal solutions (for evaluation purposes only), we must brute force the characteristic function. This is expensive and only possible for small, simple VRPs and 3 agents.

\subsection{Experimental design} We perform 10 independent runs with different random seeds to train our agents. Agents are trained for 10,000 epochs and evaluated every 100 epochs. Agents are evaluated on instances it has never seen before in training. We train using a batch size of 2048 and evaluate with a batch size of 2048. All agents use a discount factor $\gamma$ of 0.95. All agents' observations are normalised with a running estimate of the mean and standard deviation. The maximum number of bargaining rounds $T$ is set to 10. The learning rate was held constant at $3\mathrm{\times}{10^{-4}}$ and we use Adam optimisation. We clip the global norm of gradient updates if they exceed 1. We use $\varepsilon = 0.05$ to clip the probability ratios in PPO as it seems to help stability in \citep{yu_surprising_2021}. All code to generate results is run on the Wilkes 3 high performance computing cluster with AMD EPYC\texttrademark\ 7763 64-Core Processors and NVIDIA A100 GPUs. Note we only use a supercomputer to perform runs in parallel. Training takes approximately 8 hours per run. 

\subsection{Evaluation}
\subsubsection{Correlation with the Shapley value}

The objective of our work is to find a partition of the $N$ carriers with an associated fair pay-off vector. We emphasise that certain cooperative solution concepts (e.g. Shapley values) can be retrieved as the outcome of non-cooperative, extensive form games (e.g. coalitional bargaining as in our work). The Shapley value is the most common gain sharing mechanism used in the collaborative vehicle routing setting \citep{guajardo_review_2016} as it is widely accepted in game theory to be fair \--- each agent gets paid proportional to their marginal contribution. In addition, it is also guaranteed to be unique. We believe that both of these arguments would help transportation planners to reach agreements better, in line with \citep{krajewska_horizontal_2008}. Thus, we compare the outcomes that our MARL agents agree to with the Shapley value for each instance by measuring the correlation, mean absolute error, and mean squared error.

\subsubsection{Baseline bots}
\label{sec: baseline_bots}

We compare our MARL agents against two rule-based bots as a baseline. The heuristic bot always proposes the grand coalition with equal gain share and always accepts every proposal. The random bot proposes coalitions and gain shares as well as responses all uniformly at random. These two bots help us to understand that (a) our MARL agents are learning interesting, complex behaviours, and (b) our experimental setup is not too easy in design and that simple, intuitive policies are not sufficient for this setting.

\subsubsection{Accuracy} A simple evaluation metric is to measure how often the agents propose the correct coalition. For player $i$, the correct coalition $C_i^*$ is defined to be the coalition $C$ which would maximise player $i$'s reward. This involves brute forcing the characteristic function to evaluate the value of each possible coalition which is only possible since we consider 3 agents. We emphasise that brute force is only required to \emph{evaluate} our agents \--- brute force is not required to train the agents. The reward $R$ is the collaboration gain from agreeing to coalition $C$, $v(C)$, multiplied by the $i$th element of the pay-off vector, $x_i$.

\subsubsection{Optimality gap}
We denote the absolute and relative optimality gap of player $i$ by $\phi_i$ and $\eta_i$ respectively. The absolute optimality gap $\phi_i$ for player $i$ is defined as $\phi_i = v(C_i^*) - v(C)$, where $C_i^*$ is the correct coalition, $C_i$ is player $i$'s proposed coalition, and $v(\cdot)$ is the characteristic function (i.e. the collaboration gain of a given coalition). The relative optimality gap $\eta_i$, is calculated as:

\begin{equation}
  \eta_i = \frac{v(C_i^*) - v(C_i)}{v(C_i^*)}
\end{equation}

Since the data is randomly generated, there could be scenarios where there is no value in collaborating, i.e. even the value of the grand coalition is 0, $v(N) = 0$. Note that we exclude these scenarios when calculating the above evaluation metrics; however, this only occurs 1.9\% of the time when brute-forcing 51,200 instances.

\subsubsection{Other checks}
\citet{okada_noncooperative_1996} analyses this coalitional bargaining game in a non-collaborative routing setting, and proves that agents will cooperate by sharing gains equally  in our setting. Therefore, in addition to the above metrics, we check that the agents' behaviour agrees with those predicted by \citet{okada_noncooperative_1996}. Firstly, we check that agents do converge to an equal gain share. Secondly, all agents should reach agreement in the first time-step in the three-player setting.

\subsection{Results}

\begin{figure}[t]
\centering
\subfloat[Average accuracy.]{%
\resizebox*{0.60\linewidth}{!}{\includegraphics{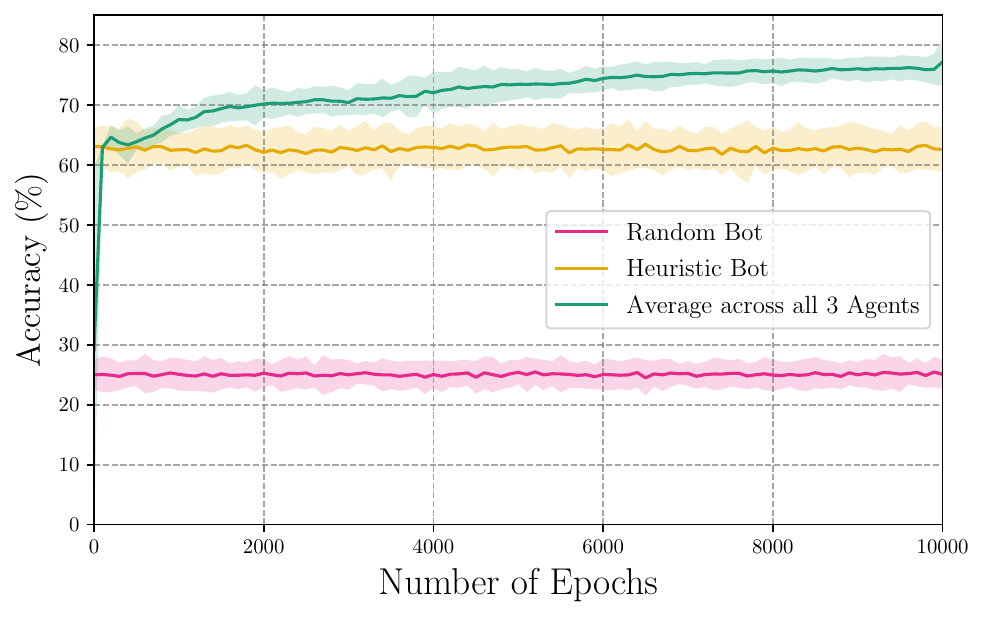}}\label{fig: accuracy}}\hspace{5pt}
\subfloat[Average optimality gap.]{%
\resizebox*{0.60\linewidth}{!}{\includegraphics{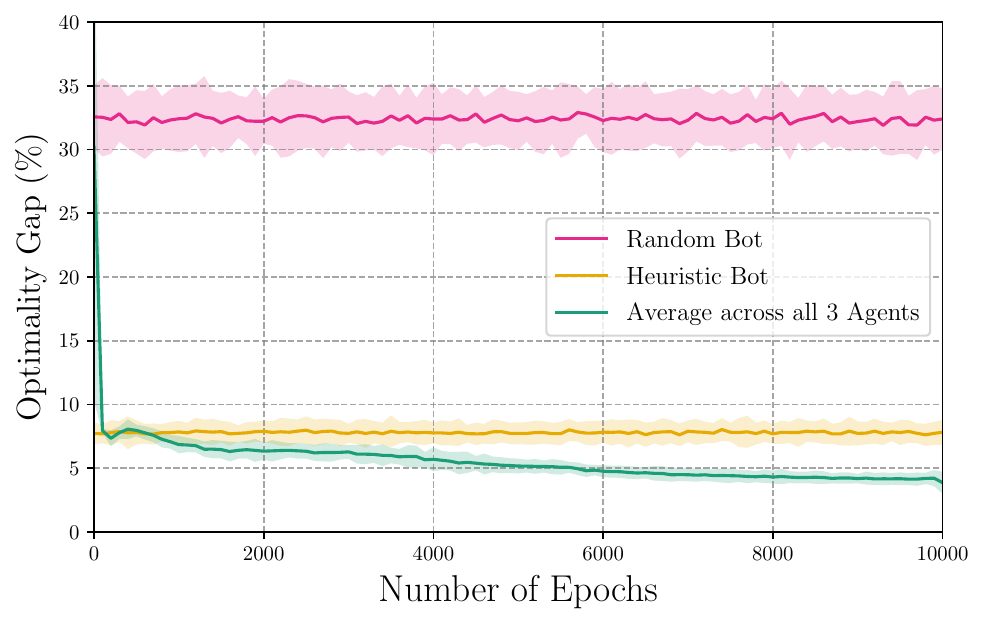}}\label{fig: optimality}}
\caption{Learning curve of (a) average accuracy (b) average optimality gap across all 3 agents for readability. Solid lines denote mean accuracy across 10 independent runs. Shaded regions denote $\pm$ two standard deviations. After training for 10,000 epochs, our RL agents reach an average accuracy of 77\% with an average optimality gap of 3.9\%.} \label{fig: f6}
\end{figure}

\begin{figure}[t]
    \centering
    \subfloat[Agent 1's proposed pay-offs.]{%
    \resizebox*{0.60\linewidth}{!}{\includegraphics{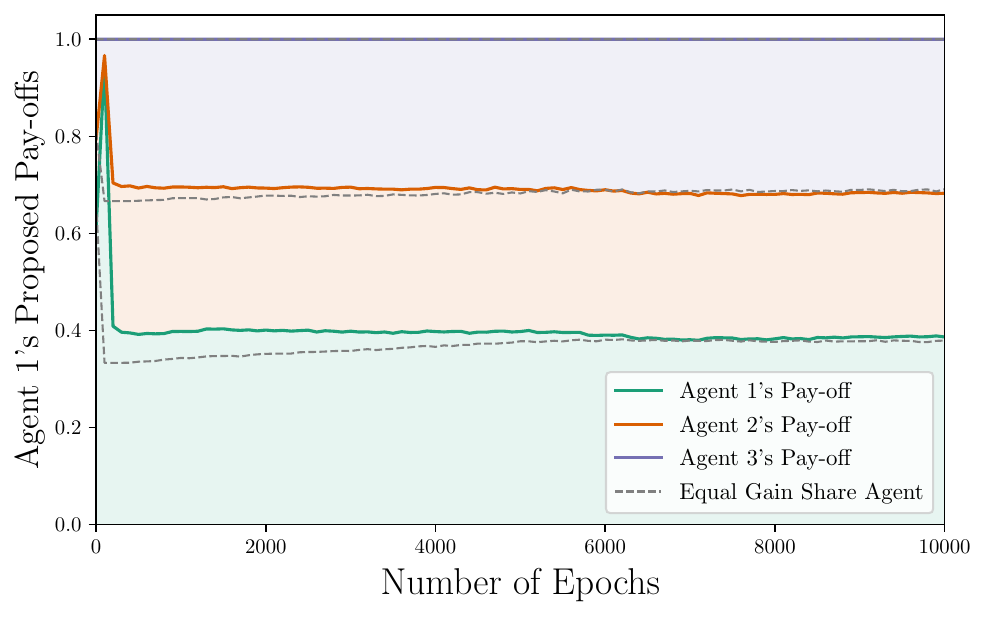}}\label{fig: payoffs}}\hspace{5pt}
    \subfloat[Average number of bargaining rounds.]{%
    \resizebox*{0.60\linewidth}{!}{\includegraphics{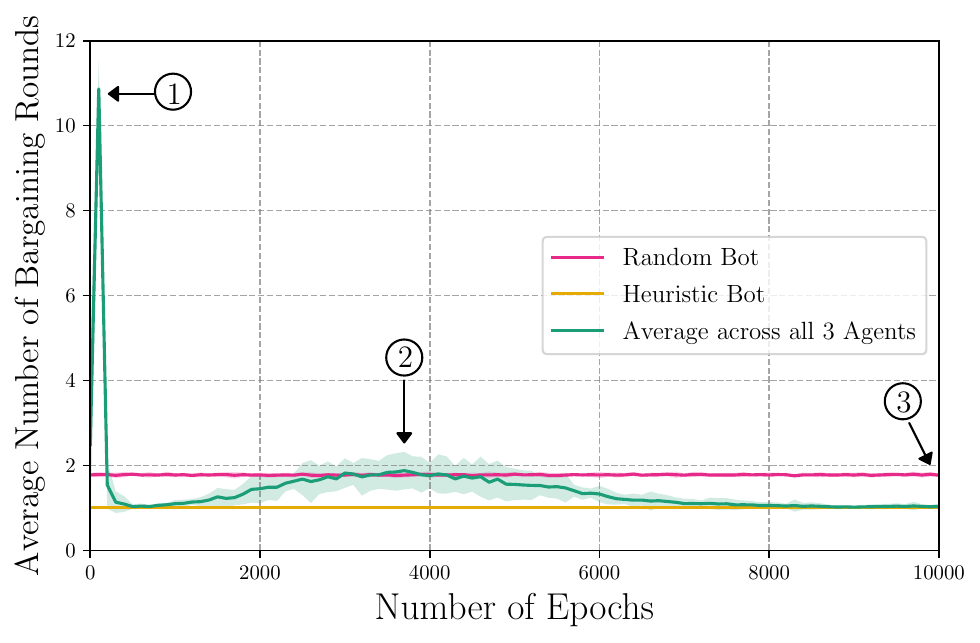}}\label{fig: round_mean}}
    \caption{Learning curve of (a) Agent 1's average proposed pay-offs (b) average number of bargaining rounds across all 3 agents for readability. Solid lines denote mean accuracy across 10 independent runs. Shaded regions denote $\pm$ two standard deviations. Dashed lines denote the proposed pay-off of an equal gain share agent. In (a), after 10,000 epochs, Agent 1 converges on an approximately equal gain share. In (b), after 10,000 epochs agents reach agreement after an averaged 1.03 rounds of bargaining. Both of these results agree with \citet{okada_noncooperative_1996}.}
    \label{sample-figure}
\end{figure}

We perform ten independent runs comparing our RL bot to the heuristic bot. Ten independent runs in (MA)RL is commonly accepted following the work of \citet{henderson_deep_2019}. We also compare to a random bot which simply proposes coalition structures and pay-off vectors as well as responds all uniformly at random. 

From Figures \ref{fig: accuracy} and \ref{fig: optimality}, we conclude that our agents have learnt close to optimal behaviour. Our agents reach an average accuracy of 77\% and average optimality gap of 0.01 (or 3.9\%). Moreover, we can see from \autoref{fig: payoffs} that Agent 1 learns to share gains equally \--- as expected by game theory \citep{okada_noncooperative_1996}. Whilst we only show the plot for Agent 1, similar plots can be made for Agents 2 and 3 but are omitted due to space constraints. Interestingly, three `phases' of learning are identified as shown in \autoref{fig: round_mean}. In Phase 1 (the first approximately 300 epochs), proposers act extremely myopically and propose that they receive the majority of the gain (up to 90\%). Occasionally, the responders will accept these sub-optimal proposals and thus the proposer could receive high reward. However, the responders learn to reject more proposals so that they can potentially counter-offer in the next round. This leads to more rounds of bargaining. After about 300 epochs, both proposers and responders reach agreement quickly; however, the gains are not equally shared. In Phase 2, responders realise they can do better by rejecting proposals and potentially proposing counter-proposals. This drives the proposers to propose more equal gain shares. Finally, in Phase 3, we can see that agents have learnt to maximally cooperate with equal gain share and reach agreement within the first time-step as expected by \citet{okada_noncooperative_1996}.

\subsubsection{Correlation with Shapley Values}

\begin{figure}[h]
    \centering
    \includegraphics[width=0.7\linewidth]{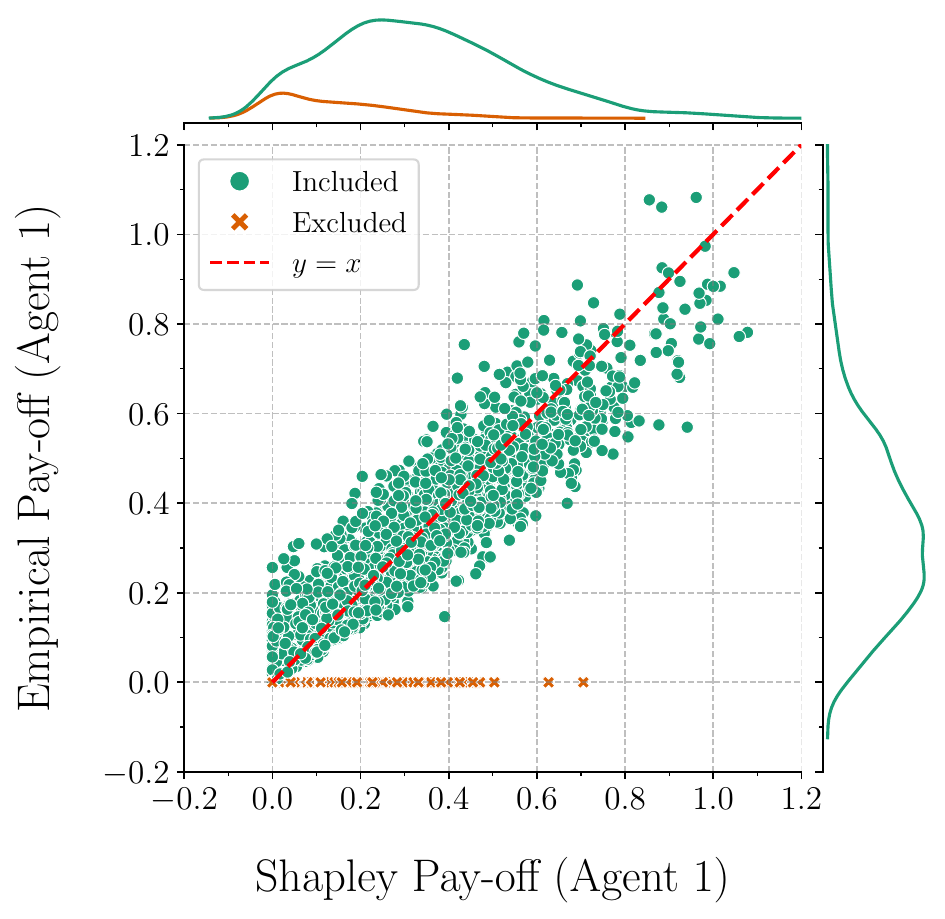}
    \caption{The empirical pay-off agent 1 receives as a result of coalitional bargaining vs. the theoretical Shapley values for 2048 test instances. Green circle markers denote when agent 1 was included in the coalition. Orange cross markers denote when agent 1 was excluded from the coalition. $R^2$ score of 0.76, mean squared error of 0.08 and mean absolute error of 0.01.}
    \label{fig: shapley}
\end{figure}

In \autoref{fig: shapley}, we see that the outcomes from our bargaining procedure correlate well with the calculated Shapley values. The three agents receive an $R^2$ score of 0.76, mean squared error of 0.08, and mean absolute error of 0.01 (averaged across all three agents). In addition, it is promising that when agent 1 is excluded from the coalition (denoted by orange cross markers), this is usually when Agent 1 has low marginal contribution (as seen by the orange kernel density estimate plot at the top of the x-axis). As a result, we conclude that our agents learn to agree to \emph{fair} outcomes. This is important from a managerial perspective as fairness could be crucial to help incentivise carriers to participate in collaborative vehicle routing \citep{guajardo_review_2016}.

\subsubsection{Ablations}
We further perform two ablations to strengthen the confidence in our findings. Each ablation is carried out with 10 random seeds each. The first ablation changes the maximum number of bargaining rounds from 10 to 30. This ablation is carried out since the underlying coalitional bargaining game is infinite-horizon, yet we must set a maximum number of bargaining rounds. Our ablation shows that increasing the maximum number of time-steps does not significantly change the quality of our agents' solutions. The agents still agree to share gains equally, with an average optimality gap of 4.1\% (up from 3.9\%) and identifying the correct coalitions 76\% of the time (down from 77\%). Therefore, we conclude that using a maximum number of time-steps of 10 to be sufficient. This is expected as we deal with time-limits properly as discussed in section \autoref{sec: time_limits}. The second ablation changes the agents' discount factor from 0.95 to 1.0. This ablation is carried out as we use a discount factor to reduce variance in the return. We test whether it's possible to use a higher discount factor. We find that using a discount factor of 1.0 decreases performance which we suspect to be due to the increased variance. With a discount factor of 1.0, agents achieve an average optimality gap of 6.07\% (up from 3.9\%). Agents do still learn to propose an approximately equal gain share but identifies the correct coalitions only 68\% of the time (down from 77\%). We conclude that using a discount factor of 0.95 is sufficient to achieve a set of strong agents.

\subsection{Discussion}
In addition, our RL agents are able to reach agreement in 512 parallel instances within an average of 3.0s (or 0.006s per instance). We note that the prior literature assumes full access to the characteristic function, such as \citet{krajewska_horizontal_2008}. Using these prior methods to solve 512 instances takes 24.3s (or 0.047s per instance). Thus, our RL agents achieve a 88\% reduction in computational time when compared with prior methods to calculate the Shapley value, such as in \citep{krajewska_horizontal_2008}. Whilst 0.047s per instance may seem reasonable even with traditional methods, we stress that this is due to the simplistic VRP setting we consider \--- prior methods will not scale with the number of agents nor problem complexity via additional constraints such as time-windows. Importantly, our agents agree to outcomes that correlate well with Shapley values and thus we conclude that our method produces \emph{fair} outcomes. This is important to fairly compensate carriers to enable wide-spread industrial adoption of collaborative vehicle routing. Our agents also reach agreement in a decentralised and self-interested manner, which overcomes the limitations of central orchestration methods mentioned in \autoref{sec: related_work}.

Furthermore, our MARL agents are able to outperform the two baseline bots in both accuracy and optimality. The heuristic bot and random bot has an accuracy of 62\% and 25\% respectively, and an optimality gap of 8\% and 32\% respectively. The relatively low performance of both the heuristic bot and random bot suggests that the experimental setup is sufficiently challenging (due to the NP-hard nature of vehicle routing problems), and that simple policies are not performant in this setting. The heuristic bot shows that 38\% of the time, it is not desirable to form the grand coalition as some agents may contribute very little. The random bot's high optimality gap shows that, whilst there is symmetry in our problem and depots are equidistant, the choice of partners is still important. This necessitates more intelligent agents and thus complex methods such as MARL. More importantly, we conclude that our MARL agents have learnt interesting behaviours, such as to exclude opponents if they contribute little to the coalition, as seen in \autoref{fig: shapley}.

In this work, we make the assumption that each carrier possesses only one truck. We further assume that the same truck driver is assigned to the same truck. This is a reasonable assumption as the road freight industry is highly fragmented: for example, in the UK, there are 60,000 registered carriers \citep{office_for_national_statistics_uk_2022} in 2022, and 1 million registered carriers in the EU in 2020 \citep{eurostat_annual_2020}. However, if a single carrier possesses multiple trucks and thus multiple drivers, it would be possible to decompose the problem at different levels of granularity. One could consider coalitions of carriers; coalitions of trucks; or even coalitions of truck drivers. Our framework should be applicable to deal with all three types of modelling choices, but clearly the more granular the modelling choice, the more computational power that will be required.

The benefit of studying collaborative routing in a coalitional bargaining game is that game theory describes optimal, rational behaviour in this setting. As a result, we have a measure of the gap to optimality. This is important because of the challenging nature of 3-player, mixed-motive settings for MARL; thus, we can understand if the agents are learning correctly. However, there are three main limitations of this approach. Firstly, collaborative vehicle routing is most fruitful with a large number of participating carriers \citep{cruijssen_joint_2007, los_large-scale_2022}. Future work must investigate scaling our MARL approach to a larger number of carriers. We believe this to be possible in a hybrid centralised-decentralised manner. The advantage of our decentralised MARL approach is that it enables us to provide a large volume of high-quality solutions to optimise the central agent. Secondly, future work should investigate the performance of MARL-based approaches on real-world data distributions with real-world constraints. One direction would be to study the effect of data imbalance (such as the locations of depots and customers, as well as the delivery volumes) on the performance of MARL-based methods. Another direction could be to study the effect of partial observability of other carriers' information; we currently consider the perfect information scenario where all delivery information is publicly shared (though, crucially, the characteristic function is still unknown). It would be interesting in future work to explore imperfect information settings, such as the value of information sharing. This could be tackled using decentralised, partially observable Markov decision processes (dec-POMDPs) \citep{oliehoek_concise_2016}. Thirdly, our approach currently only incentivises carriers. An independent third-party logistics provider may be required to enable collaborative routing. How should we incentivise third-party logistics providers? How should we incentivise shippers? What role could government play to incentivise collaboration? Moving in these directions with MARL would result in using more complex and flexible games; however, optimal, rational behaviour would be unknown. Nevertheless, MARL may still be applied to these complex games but in a descriptive manner \citep{shoham_if_2007}, i.e. to analyse the emergent behaviour of agents assuming a given MARL algorithm. We believe this to be an exciting line of future research.

\section{Conclusions and Managerial Implications}
\label{sec: conclusion}

Collaborative Vehicle Routing has promised cost savings between 4 \-- 46\% in the last two decades. Yet industrial adoption remains limited. A key remaining barrier is the design of a gain sharing mechanism that is fair and scalable such that carriers are incentivsed to collaborate. Orchestration of truck sharing is usually proposed via a central optimiser, where an intermediary would receive information from each carrier and allocate trucks to each route. Subscription to intermediaries do not necessarily outweigh costs, and carriers typically do not obtain any benefits from sharing their trucks.  In this paper, we propose an automated, decentralised approach, where software agents representing carriers find optimal routes through a \textit{coalitional bargaining} game, and any gain obtained via improved truck utilisation is shared between the carriers. Manual orchestration costs are also avoided as the approach is automated.

To facilitate decentralised optimisation and fair gain sharing  we utilised deep multi-agent reinforcement learning. The main challenge of our setting is the inability of extant methods to fully evaluate the characteristic function due to high computational complexity. The characteristic function calculates the collaboration gain for every possible coalition, which requires solving an exponential number of NP-hard VRPs. The autonomous agents designed in this work are able to correctly reason over a high-dimensional graph input to \emph{implicitly} reason about the characteristic function instead. This eliminates the need to evaluate the expensive post-collaboration vehicle routing problem an exponential number of times and increases its practicability as we only need to evaluate this once. Furthermore, applying MARL to mixed-motive games is \emph{highly} non-trivial and applying out-of-the-box MARL algorithms to this problem does not work. We show that we are able to achieve strong performance through careful design decisions, such as transfer learning, a targeted training design and COMA, and provide intuition for why these approaches help.

Moreover, the multi-agent reinforcement learning approach designed in our work is applicable to any coalitional bargaining game. Thus, our work may be suitable to problems in the broader collaborative logistics literature such as warehouse sharing. Another important point is that collaboration is not centrally orchestrated but facilitated using decentralised decision making. This marks an important step towards real-world adoption which might encourage transportation planners to consider more profitable and fair collaboration scenarios. Whilst we initially envisage this system operating as a decision support system, as transportation planners gain trust in the agents' decisions, we ultimately envisage this system to operate fully autonomously. This would enable even faster decision making that is traceable and consistent, potentially enabling a more responsive supply chain \citep{brintrup_roadmap_2009}. We urge transport planners and software system providers to consider potential adoption scenarios and integration into information systems.

Our work has limitations which provide avenues for future research. The current focus of this work is to obtain strong autonomous agents that maximally cooperate in the challenging mixed-motive setting of collaborative vehicle routing. Whilst we have achieved this, we have focused on a setting with 3 carriers as the focus of our work was to provide the theoretical link between collaborative vehicle routing, coalitional bargaining, and deep multi-agent reinforcement learning. Future work should investigate the scalability of a MARL approach to a larger number of agents. Furthermore, CVR problems typically include various additional considerations such as axle weights, goods compatibility, and packing orders, which have not yet been incorporated to the framework proposed here. Our approach is agnostic to the underlying optimisation design, and being so, we do not envisage the incorporation of additional problem features to hinder its function.

\section*{Acknowledgement(s)}
This work was performed using resources provided by the Cambridge Service for Data Driven Discovery (CSD3) operated by the University of Cambridge Research Computing Service (\url{www.csd3.cam.ac.uk}), provided by Dell EMC and Intel using Tier-2 funding from the Engineering and Physical Sciences Research Council (capital grant EP/T022159/1), and DiRAC funding from the Science and Technology Facilities Council (\url{www.dirac.ac.uk}).

We thank the three anonymous reviewers for their support and insightful comments during the review process which has greatly enhanced this paper. We also thank the Supply Chain Artificial Intelligence Lab (SCAIL) for their insightful discussions regarding early drafts of this paper.

\section*{Disclosure statement}
The authors report no conflict of interest.

\section*{Funding}
This work was supported by the UK Engineering and Physical Sciences Research Council (EPSRC) grant on ``Intelligent Systems for Supply Chain Automation'' under Grant Number 2275316, as well as by the UK EPSRC Connected Everything Network Plus under Grant EP/S036113/1.

\clearpage

\appendix

\section{Capacitated vehicle routing problem}
\label{app: b}

In our paper, the pre-collaboration social welfare can be calculated by first solving three independent Capacitated Vehicle Routing Problems, where we assume an arbitrarily high capacity for each vehicle.

The capacitated vehicle routing problem (CVRP) and their variants have been studied for over 60 years \citep{toth_vehicle_2014}. Here we show the \emph{three-index (vehicle-flow) formulation}.

The CVRP considers the setting where goods are distributed to $n$ customers. The goods are initially located at the \emph{depot}, denoted by nodes (or vertices) $o$ and $d$. Node $o$ refers to the starting point of a route, and node $d$ the end point of a route. The customers are denoted by the set of nodes $N = \{1, 2, \dots, n\}$. Each customer $i \in N$ has a \emph{demand} $q_i \geq 0$. In our setting, we consider $q_i = 1$ for all customers. A \emph{fleet} of $|K|$ vehicles $K = \{1, 2, \dots, |K|\}$ are said to be \emph{homogeneous} if they all have the same capacity $Q > 0$. In our setting, we consider only one vehicle and set its capacity $Q$ to an arbitrarily high number to remove the capacity constraint. A vehicle must start at the depot, and can deliver to a set of customers $S \subseteq N$ before returning to the depot. The \emph{travel cost} $c_{i, j}$ is associated for a vehicle travelling between nodes $i$ and $j$ which we assume to be the Euclidean distance.

This problem can be modelled as a complete directed graph $G = (V, A)$, where the vertex set $V \coloneqq N \cup \{o, d\}$ and the arc set $A \coloneqq (V \setminus \{d\}) \times (V \setminus \{o\})$. We define the \emph{in-arcs} of $S$ as $\delta^-(S) = \{(i, j) \in A: i \notin S, j \in S\}$. The \emph{out-arcs} of $S$ is $\delta^+(S) = \{(i, j) \in A: i \in S, j \notin S\}$.

The binary decision variables $x_{ijk}$ denotes whether a vehicle $k \in K$ travels over the arc $(i, j) \in A$. The binary decision variables $y_{ik}$ denotes whether a vehicle $k \in K$ visits node $i \in V$. $u_{ik}$ denotes the load in vehicle $k$ before visiting node $i$. We define the demand at the depot nodes $o$ and $d$ to be 0, i.e. $q_o = q_d = 0$. This yields:

\begin{mini!}
  {}            {\sum_{k \in K} c^T x_k \label{eq: e1o1}}                     {}               {}\tag{1a}
  \addConstraint{\sum_{k \in K} y_{ik}}                      {=1, \qquad \label{eq: e1c1}}            {\forall i \in N} \tag{1b}
  \addConstraint{x_k(\delta^+(i)) - x_k(\delta^-(i))}        {=\begin{cases} 1, & i = o, \\ 0, & i \in N, \end{cases} \qquad \label{eq: e1c2}}            {\forall i \in V \setminus \{d\}, k \in K} \tag{1c}
  \addConstraint{y_{ik}}{= x_k(\delta^+(i)) \label{eq: e1c3}}{\forall i \in V \setminus \{d\}, k \in K} \tag{1d}
  \addConstraint{y_{dk}}{= x_k(\delta^-(d)) \label{eq: e1c4}}{\forall k \in K} \tag{1e}
  \addConstraint{u_{ik} - u_{jk} + Qx_{ijk}}{\leq Q - q_j \label{eq: e1c5}}{\forall (i, j) \in A, k \in K} \tag{1f}
  \addConstraint{q_i}{\leq u_{ik} \leq Q \label{eq: e1c6}}{\forall i \in V, k \in K} \tag{1g}
  \addConstraint{x}{=(x_k) \in \{0, 1\}^{K \times A} \label{eq: e1c7}}                         {} \tag{1h}
  \addConstraint{y}{=(y_k) \in \{0, 1\}^{K \times V} \label{eq: e1c8}.}                         {} \tag{1i}
\end{mini!}

\begin{itemize}
  \item The objective function (\ref{eq: e1o1}) minimises the Euclidean distance travelled by the vehicle.
  \item Constraint (\ref{eq: e1c1}) ensures the vehicle only visits each customer once.
  \item Constraint (\ref{eq: e1c2}) ensures that the sum of vehicles entering node $d$ and exiting node $d$ is $-1$. This ensures that a vehicle $k$ performs a route starting at $o$ and ending at $d$.
  \item Constraint (\ref{eq: e1c3} and \ref{eq: e1c4}) couples variables $x_{ijk}$ and $y_{ik}$.
  \item Constraint (\ref{eq: e1c5}) is the Miller-Tucker-Zemlin constraint which helps eliminate subtours.
  \item Constraint (\ref{eq: e1c6}) is the capacity constraint.
\end{itemize}

\section{Multi-depot vehicle routing problem}
\label{app: c}

In our paper, the post-collaboration social welfare can be calculated by solving the multi-depot vehicle routing problem (MDVRP) once. The number of depots corresponds to the number of agents within the accepted coalition. Again, we remove capacity constraints by setting the capacity of each vehicle to an arbitrarily large number. However, we add the additional constraint that each vehicle has to visit at least one customer.

The MDVRP is a simple extension of the CVRP formulation provided in \autoref{app: b}. Instead of having the depot simply represented by nodes $o$ and $d$, the depots are extended to belong to a specific vehicle $k$ through nodes $o_k$ and $d_k$. Doing so yields:

\begin{mini!}
  {}            {\sum_{k \in K} c^T x_k \label{eq: e2o1}}                     {}               {} \tag{2a}
  \addConstraint{\sum_{k \in K} y_{ik}}                      {=1, \qquad \label{eq: e2c1}}            {\forall i \in V} \tag{2b}
  \addConstraint{x_k(\delta^+(i)) - x_k(\delta^-(i))}        {=\begin{cases} 1, & i = o_k, \\ 0, & i \in N, \end{cases} \qquad \label{eq: e2c2}}            {\forall i \in V \setminus \{d_k\}, k \in K} \tag{2c}
  \addConstraint{y_{ik}}{= x_k(\delta^+(i)) \label{eq: e2c3}}{\forall i \in V \setminus \{d_k\}, k \in K} \tag{2d}
  \addConstraint{y_{d_{k}k}}{= x_k(\delta^-(d_k)) \label{eq: e2c4}}{\forall k \in K} \tag{2e}
  \addConstraint{y_{d_{k}k}}{= 1 \label{eq: e2c5}}{\forall k \in K} \tag{2f}
  \addConstraint{u_{ik} - u_{jk} + Qx_{ijk}}{\leq Q - q_j \label{eq: e2c6}}{\forall (i, j) \in A, k \in K} \tag{2g}
  \addConstraint{q_i}{\leq u_{ik} \leq Q \label{eq: e2c7}}{\forall i \in V, k \in K} \tag{2h}
  \addConstraint{x}{=(x_k) \in \{0, 1\}^{K \times A} \label{eq: e2c8}}                         {} \tag{2i}
  \addConstraint{y}{=(y_k) \in \{0, 1\}^{K \times V} \label{eq: e2c9}.}                         {} \tag{2j}
\end{mini!}

\begin{itemize}
  \item The objective function (\ref{eq: e2o1}) minimises the Euclidean distance travelled by all vehicles.
  \item Constraint (\ref{eq: e2c1}) ensures that each vehicle only visits each customer once.
  \item Constraint (\ref{eq: e2c2}) ensures that the sum of vehicles entering node $d_k$ and exiting node $d_k$ is $-1$. This ensures that a vehicle $k$ performs a route starting at $o_k$ and ending at $d_k$.
  \item Constraint (\ref{eq: e2c3} and \ref{eq: e2c4}) couples variables $x_{ijk}$ and $y_{ik}$.
  \item Constraint \ref{eq: e2c5} ensures that each vehicle performs at least one delivery.
  \item Constraint (\ref{eq: e2c6}) is the Miller-Tucker-Zemlin constraint which helps eliminate subtours.
  \item Constraint (\ref{eq: e2c7}) is the capacity constraint.
\end{itemize}

\section{Expected Number of Bargaining Rounds by a Random bot}

Let $X$ be a discrete random variable denoting the number of bargaining rounds. Let's assume we have a random agent as discussed in \autoref{sec: baseline_bots} which proposes coalitions, pay-off vectors and responses uniformly at random. We wish to calculate the expected number of bargaining rounds achieved by three random bots, $\mathbb{E}[X]$. The maximum number of bargaining rounds is 10 in our experiments (although our ablations show that increasing this to 30 has no meaningful difference).

\begin{align}
    \mathbb{E}[X] &= \sum_{k=1}^{10} x \cdot P(X=x) \\
    &= 1 \cdot P(X = 1) + 2 \cdot P(X = 2) + \dots + 10 \cdot P(X=10)
\end{align}

To obtain $P(X = 1)$, note that e.g. for Player 2, the random bot can propose four coalitions, $C = \{1, 2, 3\}, \{1, 2\}, \{2, 3\}$ or $\{2\}$ since Player 2 must be in the coalition $C$. If the coalition $C = \{1, 2, 3\}$ is proposed, then both Players 1 and 3 must accept for the bargaining process to terminate, which yields a probability of acceptance (and thus termination) of $\frac{1}{2}^2$.

Therefore, $P(X=1)$ can be re-written as follows:

\begin{align}
    P(X=1) &= \left[ P(|C| = 3) \times \frac{1}{2}^2 \right] + \left[ P(|C| = 2) \times \frac{1}{2} \right] + \left[ P(|C| = 1) \right] \\
           &= \left[ 0.25 \times \frac{1}{2}^2 \right] + \left[ (0.25 + 0.25) \times \frac{1}{2} \right] + \left[ 0.25 \right]
\end{align}

Repeating a similar logic to calculate $\mathbb{E}[X]$ yields an expected number of bargaining rounds of 1.775.

\begin{align}
    \mathbb{E}[X] &= (1 \cdot 0.5625) + (2 \cdot 0.2461) + (3 \cdot 0.1077) + (4 \cdot 0.0471) + \dots + (10 \cdot 0.0003)\\
    &= 1.775
\end{align}

Empirically, our bots reach agreement at 1.777 rounds averaged over 10 runs.


\section{Pseudo-code of the entire pipeline}
\label{sec: pseudo_code}

\begin{algorithm}[H]
\caption{Pseudo-code of MARL pipeline}
\begin{algorithmic}[1]
\State Initialise $\boldsymbol{\theta} = \theta_1, \theta_2, \dots, \theta_n, \theta_{critic}$  \hfill {\gray{// $n$ actors' (neural network) policy and critic}}
\\
\State \gray{{// Supervised Pre-training (regression, minimise mean-squared error)}}
\For {$\theta_i$ in $\boldsymbol{\theta}$}
    \State $(\Delta \hat{y}_{\theta_i})^2 = (v(C) - \hat{y}_{\theta_i})^2$ \hfill \gray{// Calculate loss}
    \State $\Delta \theta_i = \nabla_{\theta_i}(\Delta \hat{y}_{\theta_i})^2 $ \hfill \gray{// Calculate gradients}
    \State $\theta_{i} = \theta_{i} + \alpha \Delta \theta_i$  \hfill \gray{// Update parameters}

\EndFor
\\
\State \gray{{// MARL Training}}
\For {each training epoch $e$}
    \State Initialise $M = 2048$ parallel environments \hfill \gray{{// Coalitional bargaining envs.}}
    \State $s_1 \sim \rho_1, t = 0$ \hfill \gray{{// Sample the initial state $s_1$ from $\rho_1$}}

    \While {$s_t \neq$ terminal \textbf{and} $t < T$}
        \State t += 1

        \State \gray{// Calculate joint actions \textbf{a}}
        \For {i in N}
            \State $a_{i, t} \sim \pi_{\theta_i}(a_{i, t} | s_{i, t})$  \hfill \gray{// Select actions stochastically for exploration}
        \EndFor
        
        \State $s_{t+1} \sim \mathcal{T}(s_t, A_t)$  \hfill \gray{// Sample next state from transition dynamics}

        \State $R_{i, t} \sim \mathcal{R}(s_t, A_t, s_{t+1}) \quad \forall i \in N$ \hfill \gray{// Calculate reward}

        \State Store each $\langle s_{i, t}, a_{i, t}, \log(\pi_{\theta_i}(a_{i, t} | s_{i, t})), s_{i, t+1}, R_{i, t} \rangle \forall i \in N$ in agent $i$'s buffer

    \EndWhile
    \\
    \State \gray{{// Here, all $M$ episodes will be finished}}
    \For {$t=1$ \textbf{to} T}
    \State $G_{i, t} = \sum_{t'=t}^T \gamma^{t' - t} R_{i, t} \quad \forall i \in N$  \hfill \gray{// Calculate discounted returns}
    \EndFor
    \For {t=T \textbf{down to} 1}
    \State $(\Delta Q_{i, t})^2 = \left[ G_{i, t} - \hat{Q}_{\theta_{critic}}(s_{i, t}, \textbf{a}) \right]^2$ \hfill \gray{// Calculate critic loss}
    \State $\Delta \theta_{critic} = \nabla_{\theta_{critic}}(\Delta Q_{i, t})^2 $ \hfill \gray{// Calculate critic gradients}
    \State $\theta_{critic} = \theta_{critic} + \alpha \Delta \theta_{critic}$  \hfill \gray{// Update critic parameters}    
    \EndFor
    \For {t=T down to 1}
        \State \gray{// Calculate proposal baseline}
        \State $A_{t,\,prop.}^i = G_{i, t}  - \hat{V}(s, \theta_{critic}) \quad \forall i \in N$
        
        \vspace{0.25cm}
        
        \State \gray{// Calculate response baseline}
        \State $A_{t,\,resp.}^i = G_{i, t} - \sum_a \hat{Q}(s_{i, t}, a, \textbf{a}^{-a}, \theta_{critic}) \pi_{\theta_i}(a_{i, t} | s_{i, t})  \quad \forall i \in N$
        
        \vspace{0.25cm}
        
        \State \gray{// Accumulate actor proposal gradients}
        \State $\Delta \theta_{i} \mathrel{+}= \nabla_{\theta_i} \left[ \min(r_t(\theta_i) A_{t, prop.}^i, \text{clip}(r_t(\theta_i), 1 - \varepsilon, 1 + \varepsilon) A_{t, prop.}^i  \right] \forall i \in N$
        
        \vspace{0.25cm}
        
        \State \gray{// Accumulate actor response gradients}
        \State $\Delta \theta_{i} \mathrel{+}= \nabla_{\theta_i} \left[ \min(r_t(\theta_i) A_{t, resp.}^i, \text{clip}(r_t(\theta_i), 1 - \varepsilon, 1 + \varepsilon) A_{t, resp.}^i  \right]  \forall i \in N$

        \vspace{0.25cm}

        \EndFor
    \State $\theta_i = \theta_i + \alpha \Delta \theta_{i} \quad \forall i \in N$ \hfill \gray{// Update actors' policy parameters}
    \State $\Delta \theta_i = \boldsymbol{0}$ \hfill \gray{// Reset gradients}
    \EndFor

\end{algorithmic}
\end{algorithm}


\section{Holistic Diagram of our Pipeline}
\label{sec: holistic_diagram}

\begin{figure}[h]
    \centering
    \includegraphics[width=0.99\linewidth]{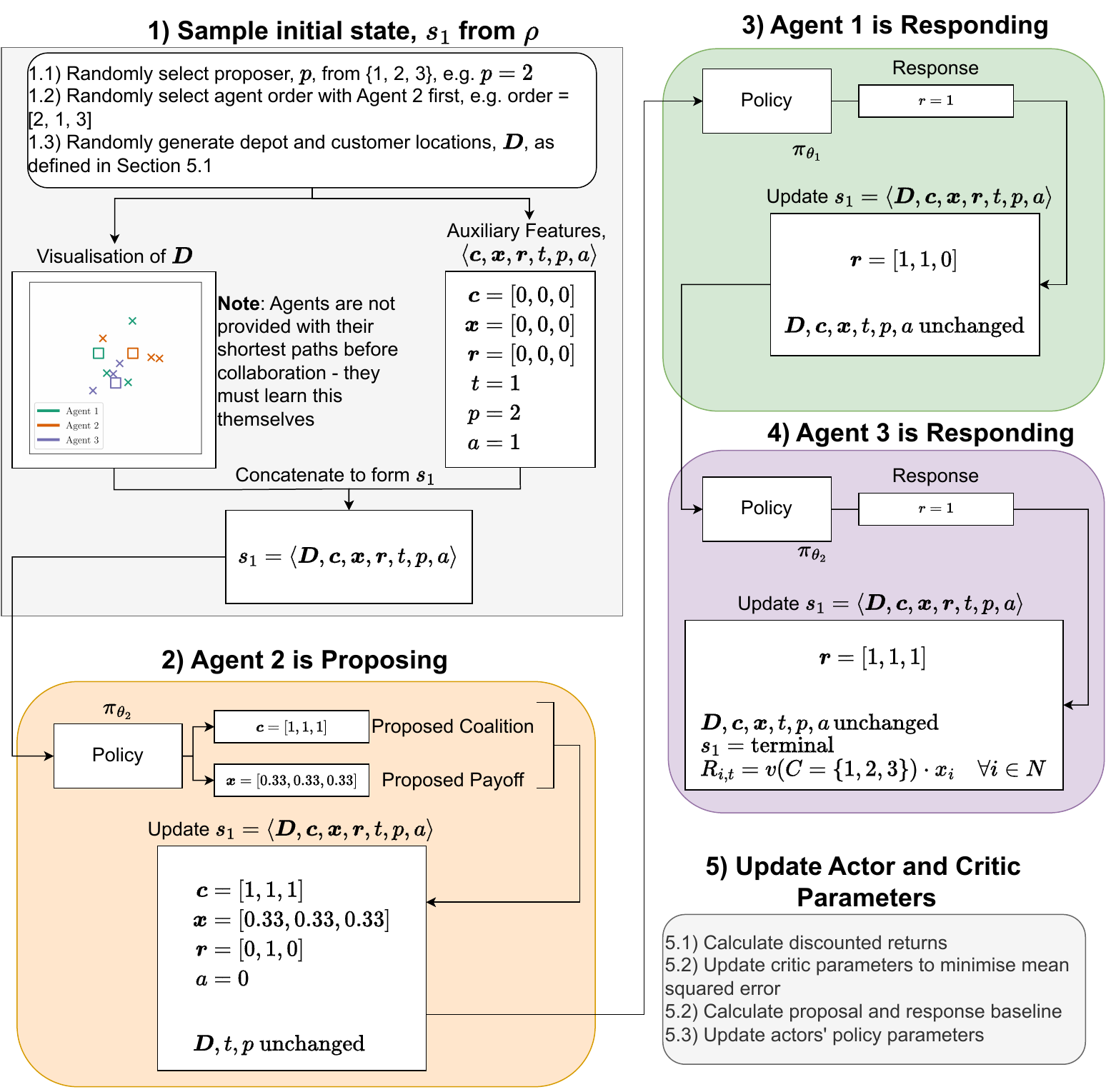}
    \caption{A holistic diagram of our proposed approach. We depict a sample trajectory where Agent 2 is selected to propose and that Agent 2 proposes to form the grand coalition with an equal payoff vector. Agents 1 and 3 then agree to the proposal. Finally, the actors' parameters and critic's parameters are updated accordingly.}
    \label{fig: holistic_diagram_full}
\end{figure}

\clearpage

\bibliography{references.bib}

\end{document}